\documentclass[conference]{IEEEtran}
\IEEEoverridecommandlockouts
\usepackage{filecontents}
\usepackage{amsmath,amssymb,amsfonts}
\usepackage{graphicx}
\usepackage{textcomp}
\usepackage{xcolor}
\usepackage{graphicx,verbatimbox,listings,caption,float,lipsum,amsmath,amssymb,algorithm,algpseudocode}
\usepackage{multirow}
\usepackage{array}
\usepackage{subcaption}
\usepackage{booktabs}

\usepackage[misc]{ifsym}
\def\BibTeX{{\rm B\kern-.05em{\sc i\kern-.025em b}\kern-.08em
    T\kern-.1667em\lower.7ex\hbox{E}\kern-.125emX}}
\begin{document}

\title{Responsible Multilingual Large Language Models: A Survey of Development, Applications, and Societal Impact}

\author{Junhua Liu\\
\textit{Forth AI}\\
j@forth.ai
\and
Bin Fu\\
\textit{Shopee}\\
bin.fu@shopee.com}

\maketitle

\begin{abstract}
Multilingual Large Language Models (MLLMs) represent a pivotal advancement in democratizing artificial intelligence across linguistic boundaries. While theoretical foundations are well-established, practical implementation guidelines remain scattered. This work bridges this gap by providing a comprehensive end-to-end framework for developing and deploying MLLMs in production environments. We make three distinctive contributions: First, we present an actionable pipeline from data pre-processing through deployment, integrating insights from academic research and industrial applications. Second, using Llama2 as a case study, we provide detailed optimization strategies for enhancing multilingual capabilities, including curriculum learning approaches for balancing high-resource and low-resource languages, tokenization strategies, and effective sampling methods. Third, we offer an interdisciplinary analysis that considers technical, linguistic, and cultural perspectives in MLLM development. Our findings reveal critical challenges in supporting linguistic diversity, with 88.38\% of world languages categorized as low-resource, affecting over a billion speakers. We examine practical solutions through real-world applications in customer service, search engines, and machine translation. By synthesizing theoretical frameworks with production-ready implementation strategies, this survey provides essential guidance for practitioners and researchers working to develop more inclusive and effective multilingual AI systems.

\end{abstract}

\vspace{.3cm}

\section{Introduction}

Multilingual large language models (MLLMs) aim to address linguistic diversity in Artificial Intelligence (AI), acknowledging that over 7,000 languages are spoken worldwide~\cite{ethnologue}, with hundreds of them having millions of speakers. 

Thanks to the recent growth with the advent of large language models (LLMs), the Natural Language Processing (NLP) research community has been enabling more advanced applications in text understanding and generation. However, much of this progress has been concentrated on high-resource languages, particularly English.

Developing truly multilingual systems remains a significant challenge, as many languages are underrepresented in current AI technologies. Research and development in this area are critical to ensuring equitable access to technological advancements for speakers of all languages.

This study addresses the lack of multilingual support and linguistic inclusivity in state-of-the-art LLMs. We present emerging trends and promising directions in the development of MLLMs that can better address linguistic diversity and representation challenges.

Our findings contrast with previous efforts that have mainly focused on a few dominant languages. We offer new insights into technical strategies that can support more equitable language processing across diverse linguistic groups.

This work highlights the importance of addressing linguistic inclusion in AI and suggests a path toward more comprehensive and inclusive NLP systems. Understanding these advancements will have broad implications for global AI accessibility and for bridging language-based technological divides~\cite{ruder_nlp}.

\subsection{Contributions}
Compared to existing MLLM surveys~\cite{xu2024survey, qin2024multilingual, huang2024survey}, this work emphasizes the practical aspects of MLLMs from development to production, based on empirical experiences. Specifically, the three primary contributions are summarised as follows:

\subsubsection{MLLM Production Framework}
We present a comprehensive end-to-end framework for developing and deploying MLLMs, bridging the gap between theoretical understanding and practical implementation. Unlike previous surveys that focus on specific aspects, our work provides an integrated view of the complete MLLM life cycle - from data collection and pre-processing through model development to industrial applications. This holistic approach offers concrete guidance for both researchers and practitioners working to develop or improve multilingual models.

\begin{table*}[t]
    \centering
    \scalebox{1.4}{
    \begin{tabular}{|c|c|c|c|c|c|}
    \hline
    \multirow{2}{*}{\textbf{Language}} & \multirow{2}{*}{\textbf{Population (2021)}} & \multirow{2}{*}{\textbf{No. of Users}} & \textbf{Internet} & \textbf{Users Growth} & \textbf{Users} \\
     &  &  & \textbf{Penetration} & \textbf{(2000 - 2021)} & \textbf{Distribution}  \\ \hline
    English & 1,531,179,460 & 1,186,451,052 & 77.5\% & 742.9\% & 25.9\% \\ \hline
    Chinese & 1,477,137,009 & 888,453,068 & 60.1\% & 2,650.4\% & 19.4\% \\ \hline
    Spanish & 516,655,091 & 363,684,939 & 70.4\% & 1,511.0\% & 7.9\% \\ \hline
    Arabic & 447,572,591 & 237,418,349 & 53.0\% & 9,348.0\% & 5.2\% \\ \hline
    Portuguese & 290,939,425 & 171,750,818 & 59.0\% & 1,948.4\% & 3.7\% \\ \hline
    Bahasa / Malay & 306,327,093 & 198,029,815 & 64.6\% & 3,356.0\% & 4.3\% \\ \hline
    French & 431,503,424 & 151,733,611 & 35.2\% & 592.7\% & 3.3\% \\ \hline
    Japanese & 126,376,461 & 116,626,672 & 92.3\% & 152.0\% & 2.5\% \\ \hline
    Russian & 145,934,462 & 116,353,942 & 79.8\% & 3,453.6\% & 2.5\% \\ \hline
    German & 92,654,451 & 92,525,427 & 99.9\% & 186.7\% & 2.0\% \\ \hline
    \textbf{Top 10} & 5,273,725,132 & 3,525,207,347 & 68.8\% & 1,188.2\% & 76.9\% \\ \hline
    \textbf{The Rest} & 2,522,890,578 & 1,060,551,371 & 42.0\% & 1,114.1\% & 23.1\% \\ \hline
    \textbf{Total} & 7,796,615,710 & 4,585,758,718 & 58.8\% & 1,170.3\% & 100.0\% \\ \hline
    \end{tabular}}
    \caption{Top Ten Languages Used on the Web - March 31, 2020.}
    \label{table:topten}
    \end{table*}
    
\subsubsection{Optimization Strategies}
We provide detailed, practical optimization strategies for multilingual models, using Llama2 as a concrete case study. We analyze specific techniques for enhancing multilingual capabilities, including curriculum learning approaches for balancing high-resource and low-resource languages, optimization of tokenization strategies, and effective sampling methods. This practical focus fills a crucial gap in existing literature, offering actionable insights for improving MLLM performance across diverse languages.

\subsubsection{Interdisciplinary Considerations}
We take an interdisciplinary approach that integrates technical, linguistic, and cultural perspectives. Our analysis extends beyond purely computational aspects to consider how linguistic diversity and cultural representation affect model development and performance. We examine how these factors influence everything from tokenization challenges to evaluation metrics, providing a more nuanced understanding of the complexities involved in creating truly effective multilingual models. This socio-technical analysis helps identify both technical and cultural barriers to achieving better multilingual performance, along with strategies for addressing them.

These contributions collectively provide a foundation for understanding and advancing the field of MLLMs, while offering practical guidance for addressing current challenges in both academic research and industrial applications.

\section{Overview of Multilingual Research}

In today's global landscape, there are over 7,000 languages in use worldwide, with more than 400 languages having over one million speakers each, and approximately 1,200 languages having over 100,000 speakers. This section will focus on examining the critical questions of whether multilingual research and language preservation are necessary, and whether it is essential to ensure that speakers of all languages can equitably benefit from the technological advances brought by large language models. Additionally, this section will provide readers with a concise overview of the development trends and technical directions in MLLMs.

\subsection{Multilingual Research}
Multilingual research in Natural Language Processing (NLP) is crucial for several reasons:

\begin{itemize}
    \item \textbf{Linguistic Diversity:} With over 7,000 languages spoken worldwide~\cite{ethnologue}, focusing solely on high-resource languages like English excludes a vast majority of the world's linguistic diversity.
    
    \item \textbf{Digital Inclusion:} As shown in Table~\ref{table:topten}, there's a significant disparity in internet penetration and usage across languages. Multilingual NLP can help bridge this digital divide and provide equal access to information and technology.
    
    \item \textbf{Cultural Preservation:} Many low-resource languages are at risk of digital extinction. Developing NLP tools for these languages can aid in their preservation and revitalization.
    
    \item \textbf{Economic Opportunity:} Enabling NLP capabilities in more languages can open up new markets and economic opportunities for speakers of those languages.
    
    \item \textbf{Fairness and Representation:} Focusing on multiple languages helps reduce bias in AI systems and ensures fair representation of diverse linguistic communities.
\end{itemize}

Despite the clear importance, developing truly multilingual NLP systems remains challenging due to resource constraints, linguistic complexity, and technical limitations. Addressing these challenges is essential for creating more inclusive and globally accessible AI technologies.

We will discuss the importance of multilingual research from four aspects:

\subsubsection{Social Aspect}
The language people use determines the education, knowledge, and social networks they can access. Although the internet is open, there still exists a Digital Language Divide between mainstream languages (Chinese, English, and other major Western languages) and other languages. As shown in Table~\ref{table:topten}, only a few languages appear frequently on the internet, which greatly increases the barrier for speakers of minority languages to access information.

The existence of the digital language divide affects the development of NLP technology at various levels. For example, most social apps (like WeChat) currently use informal input forms (such as abbreviations or slang) for commonly used languages. However, the digital language divide means that speakers of non-mainstream languages cannot get good keyboard input support and spelling correction services. This sometimes leads to bias and discrimination in NLP algorithms against non-English users~\cite{qz2017google}, which may not be caused by the language itself, but by different accents affecting algorithm results~\cite{alexa_accent}.

This neglect of non-mainstream languages not only widens the digital language divide but may also cause some non-mainstream language speakers to abandon their languages in favor of languages with better technical support and richer resources, thus affecting linguistic diversity. Therefore, to allow speakers of non-mainstream languages to better enjoy the convenience brought by technological development, reduce the performance gap of algorithms for different languages, and eliminate language barriers, it is necessary to make models cover more languages beyond English and Chinese.

\begin{table}[t]
    \centering
    \begin{tabular}{|l|c|l|l|}
    \hline
    \textbf{Language} & \textbf{ISO} & \textbf{Language Family} & \textbf{Pretrained BERT Model} \\ \hline
    Arabic    & AR & Afroasiatic    & AraBERT~\cite{antoun2020arabert} \\ \hline
    English   & EN & Indo-European  & BERT~\cite{devlin2018bert} \\ \hline
    Finnish   & FI & Uralic         & FinBERT~\cite{virtanen2019multilingual} \\ \hline
    Indonesian & ID & Austronesian  & IndoBERT~\cite{wilie2020indonlu} \\ \hline
    Japanese  & JA & Japonic        & Japanese-char BERT~\cite{cl_to_bert_japanese}\\ \hline
    Korean    & KO & Koreanic       & KR-BERT~\cite{lee2020study} \\ \hline
    Russian   & RU & Indo-European  & RuBERT~\cite{kuratov2019adaptation} \\ \hline
    Turkish   & TR & Turkic         & BERTurk~\cite{schweter2020berturk} \\ \hline
    Chinese   & ZH & Sino-Tibetan   & Chinese BERT~\cite{devlin2018bert} \\ \hline
    \end{tabular}
    \caption{Pretrained BERT Models for Various Languages}
    \label{table:7-2}
    \end{table}

\subsubsection{Linguistic Aspect}
When training large language models, we always hope to train a language-independent, more general large model. However, most of the time, limited by corpus resources, the models we train are only proficient in single-language tasks, such as Chinese or English tasks. Table~\ref{table:7-2} lists single-language BERT models for different languages and language families. Single-language pre-trained large models in Chinese, English, and some high-resource languages cannot represent other languages in the world.

Taking Chinese as an example, it belongs to the Sino-Tibetan language family and is a morphologically poor language that focuses more on expressing meaning through syntax, i.e., emphasizing word order (such as subject-verb-object order). It needs to use different words rather than just word changes to express tense, singular/plural, and gender~\cite{zhihu2022ai}. For example, in Manipuri (Meitei), te/de at the end of a sentence is used to express negation: thak-ke translates to "I drank," while thak-de translates to "I did not drink"~\cite{chelliah2001sinotibetan}.

\begin{figure}[t]
    \centering
    \includegraphics[width=.48\textwidth]{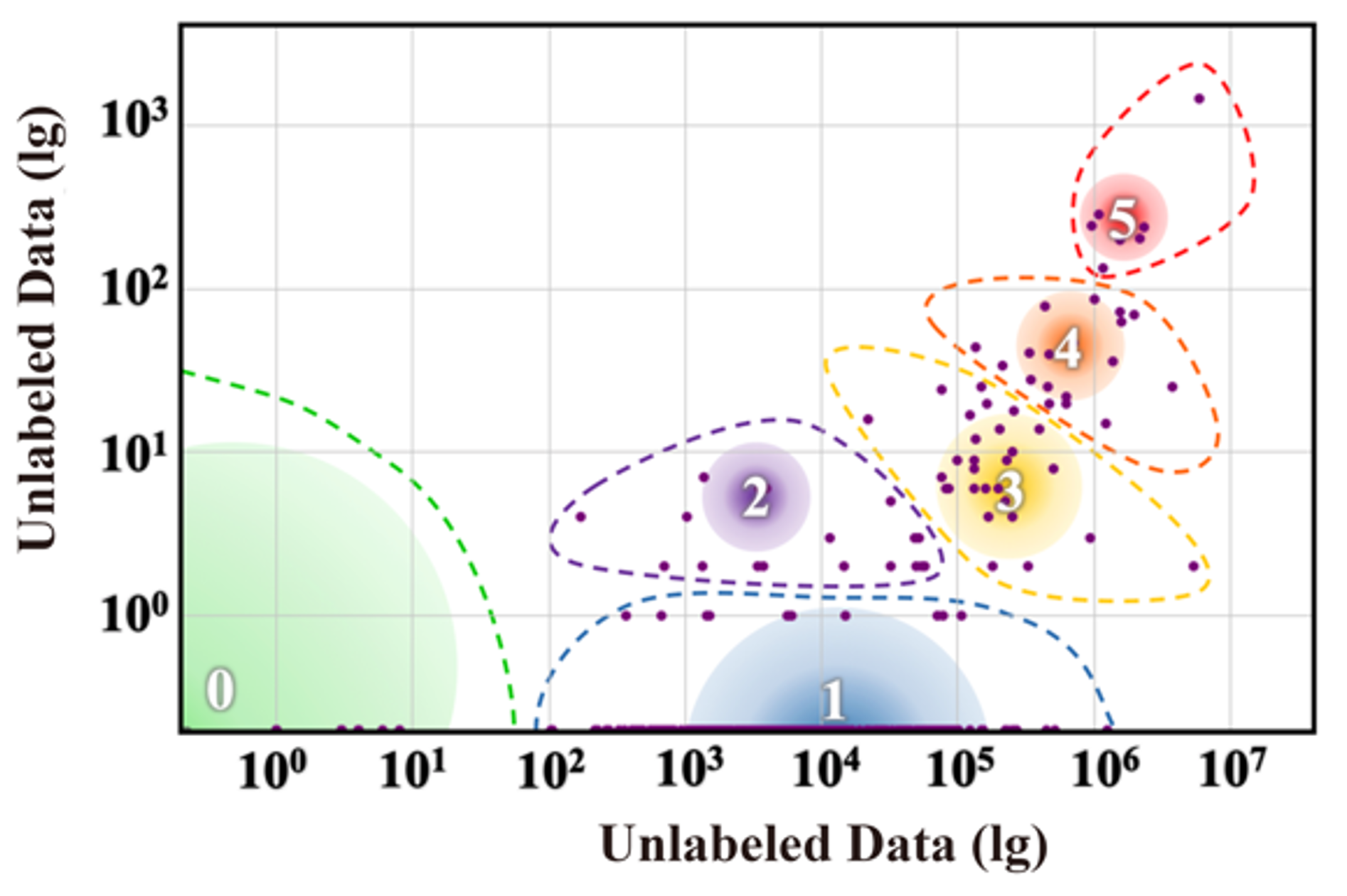}
    \caption{Different categories of language resources in NLP systems}
    \label{fig:unlabelled}
\end{figure}

Apart from these lexical and syntactic differences, we can also look at the different characteristics of various languages from a more comprehensive perspective. The World Atlas of Language Structure~\cite{wals2024} categorizes 192 features, with subject-verb-object (SVO) order being one of them. Each feature contains an average of 5.93 categories. For example, the SVO order feature can have different categories such as SOV, SVO, VSO, etc. Research has found that 48\% of the features only appear in low-resource languages, which are groups 0-2 in Figure~\ref{fig:unlabelled} of the World Atlas of Language Structure, and do not appear in groups 3-5~\cite{joshi2020state}. If we don't use these features when training or fine-tuning large models, we may miss some valuable information for model generalization.

Aiming to train better MLLMs can help us better understand the relationships between various languages in the world~\cite{artetxe2020call}, and in turn, it can help models better utilize linguistic features for downstream tasks~\cite{liu2022title2vec,liu2024spatial}. You can use your knowledge of non-native languages to explore how they differ from your native language in various aspects, such as the use of compounding, derivation, and reduplication phenomena.

\subsubsection{Cultural and Ethical Aspect}
The data we use to train large models not only reflects the characteristics of the corresponding language but also reveals cultural identity and common sense through the model's responses. However, different cultures have different common sense, such as alcohol culture not appearing in Arab countries. Some smaller large models, such as 13B or 7B models, if only fine-tuned for dialogue in one language, such as Chinese, will give reasonable answers in Chinese dialogue, but will exhibit hallucination phenomena common to large models for other languages, giving answers that are inconsistent with facts or violate ethical norms.

Large models are now being used for various complex generation tasks rather than simple classification tasks. Therefore, how to use multilingual resources to train large models that conform to the cultural and ethical norms of speakers of all languages in the world, and promote the development of Responsible AI (RAI) in different languages, has become an important topic.

\begin{table*}[t]
    \centering
    \scalebox{1.2}{
    \begin{tabular}{@{}ccccc@{}}
    \toprule
    \multirow{2}{*}{Category} & Example & Number of & Population & Proportion of\\
            & Languages & Languages & (100M) & Languages (\%) \\
    \midrule
    \multirow{2}{*}{0} & Dahalo, Warlpiri, & \multirow{2}{*}{2291} & \multirow{2}{*}{12} & \multirow{2}{*}{88.38} \\
      & Popoloca, Wallisian, Bora & & & \\
    \midrule
    \multirow{2}{*}{1} & Cherokee, Fijian, & \multirow{2}{*}{222} & \multirow{2}{*}{0.3} & \multirow{2}{*}{5.49} \\
      & Greenlandic, Bhojpuri, Navajo & & & \\
    \midrule
    \multirow{2}{*}{2} & Zulu, Konkani, Lao, & \multirow{2}{*}{19} & \multirow{2}{*}{0.057} & \multirow{2}{*}{0.36} \\
      & Maltese, Irish & & & \\
    \midrule
    \multirow{2}{*}{3} & Indonesian, Ukrainian, & \multirow{2}{*}{28} & \multirow{2}{*}{18} & \multirow{2}{*}{4.42} \\
      & Cebuano, Afrikaans, Hebrew & & & \\
    \midrule
    \multirow{2}{*}{4} & Russian, Hungarian, & \multirow{2}{*}{18} & \multirow{2}{*}{22} & \multirow{2}{*}{1.07} \\
      & Vietnamese, Dutch, Korean & & & \\
    \midrule
    \multirow{2}{*}{5} & English, Spanish, German, & \multirow{2}{*}{7} & \multirow{2}{*}{25} & \multirow{2}{*}{0.28} \\
      & Japanese, French & & & \\
    \bottomrule
    \end{tabular}}
    \caption{Examples of Different Language Categories, Users, and Proportions}
    \label{tab:lang_categories}
    \end{table*}
    
\subsubsection{Model Aspect}
When we train large models, they develop inductive biases towards languages that occupy a larger proportion of the training corpus, even if we don't explicitly encode language information in neural networks and only use N-grams to build language models. Research shows that their performance also significantly decreases on morphologically rich languages~\cite{bender2011language_independence}.

Similarly, Transformer-based models also ignore the complexity of morphologically rich languages~\cite{tsarfaty2020spmrl}: subword tokenizers perform poorly on languages with reduplication~\cite{vania2017morphology}; BPE algorithms cannot align morphological information well; although large models have shown zero-shot learning ability in monolingual and cross-lingual tasks, their text generation and classification performance decline in low-resource languages or languages that are categorically distant from mainstream training languages~\cite{zeng2023soft}; compared to large models that have undergone task fine-tuning, there is still a large performance gap~\cite{muennighoff2022crosslingual}.

These issues pose challenges to how to use word and sentence information to build MLLMs. In recent years, scholars have also begun to focus on the cross-lingual learning ability of MLLMs on low-resource languages, and the in-context learning (ICL) ability of multilingual combined prompts on large models.

\subsection{Challenges in Multilingual NLP}
After discussing the necessity of multilingual research, we look at the current multilingual research communities, challenges, and research directions.

\subsubsection{Current Trends}
There are now many research institutions dedicated to multilingual research, covering both languages that cover large populations like Chinese, Japanese, Turkish, and Hindi, as well as languages that cover small populations like Irish. In recent years, some NLP communities have also appeared that specifically research under-represented languages or language families, and more NLP communities focus on regional languages. For example, Masakhane researches African languages, AmericasNLP researches Native American languages, and IndoNLP researches Indonesian languages. At the same time, there are also dedicated long-term Workshops and Events for non-English research, such as the regular conferences organized by the Chinese Information Processing Society of China (such as CCL, ACL), and special interest groups set up for linguistic typology (such as SIGTYP, AfricaNLP, ArabicNLP, and ComputeEL).

Meanwhile, some communities focus on broader languages and work, such as ML Collective and Big Science. Big Science is a community dedicated to serving multilingual AI, which released the BLOOM model~\cite{scao2022bloom} and performed multi-task prompt fine-tuning on BLOOM and mT5~\cite{xue2020mt5}, two MLLMs, to produce BLOOMZ and mT0 with strong cross-lingual capabilities~\cite{muennighoff2022crosslingual}.

To highlight the importance of multilingualism, ACL not only established SIGTYP but also set up a Special Theme Track in 2022, aiming to make scientific papers accessible to more people through the following efforts:
\begin{itemize}
    \item Translating the ACL Anthology into 60 languages.
    \item Dubbing and adding subtitles to all conferences in 10 languages.
    \item Translating a comprehensive and standard set of NLP terminology into 60 languages.
\end{itemize}

These resources and terminology lists can allow people from different regions to discuss NLP technology in their own languages.

\subsubsection{Challenges}
Here we introduce two typical multilingual development challenges:

\textbf{Curse of Multilinguality.} Why can current MLLMs cover at most over 100 languages? Apart from resource issues, another reason is the Curse of Multilinguality~\cite{conneau2019unsupervised}. Similar to training large models on multiple tasks, the more languages used to train large models, the harder it becomes for the model to learn the representation information of each language due to the limited model capacity (a few hundred MB). The emergence of MLLMs has broken this bottleneck, with parameter scales of over 10B allowing large models to better learn the representation information of each language~\cite{goyal2021larger}.

\textbf{Low-resource Problem.} A primary issue in the development of MLLMs is that the available corpus resources show a long-tail distribution. The high-resource corpora we often talk about are heavily biased towards Indo-European languages represented by English and Chinese, Japanese, Korean, etc. These head languages have enormous quantities in both labeled and unlabeled data. We have divided the world's 7000+ languages into 6 categories from 0 to 5 based on two dimensions ~\cite{joshi2020state}: labeled data and unlabeled data, respectively representing the difficulty for large models to utilize the language, with 0 being the most difficult. As shown in Fig.~\ref{fig:unlabelled}, the areas enclosed by dotted lines of different colors represent the number of languages included, with the colors from dark to light representing the number of speakers of these languages from high to low.

\textbf{Category 0 (The Left-Behinds)}:
NLP technology has consistently ignored Category 0 languages (as shown in Class 0 in Table~\ref{tab:lang_categories}. Due to extremely limited corpus resources, these languages will gradually become historical artifacts, making it difficult to connect them with digital transformation. Even using unsupervised learning methods would only make things worse, as there is essentially no unlabeled data available.

\textbf{Category 1 (The Scraping-Bys)}:
Category 1 languages (such as Greenlandic) have some unlabeled data of certain scale, making it possible for them to receive more attention from researchers in multilingual research in the coming years. However, this requires researchers to systematically and continuously promote tasks involving these languages, get more people to pay attention to them, and be willing to collect more labeled data for them. Currently, these languages have almost zero labeled data.

\textbf{Category 2 (The Hopefuls)}:
Category 2 languages (such as Irish) are experiencing darkness before dawn in the NLP field but continue to make progress. These languages have accumulated some labeled data, albeit in small quantities, indicating that there is a group of active researchers working on digital transformation of these languages. They are expected to develop some promising NLP tools for these languages in the coming years.

\textbf{Category 3 (The Rising Stars)}:
From Category 3 onwards, we're basically talking about what we commonly refer to as high-resource languages. Unsupervised learning has greatly accelerated the influence of these languages (such as Indonesian) in the NLP field. Because these languages have more users on the internet, there are some flourishing NLP community researchers dedicated to studying these languages, although their research is also affected by insufficient labeled data. These researchers should leverage large model pre-training and parameter-efficient fine-tuning techniques~\cite{peft} to compensate for the lack of labeled data.

\textbf{Category 4 (The Underdogs)}:
Category 4 languages (such as Vietnamese) are like sparks in the NLP field, with huge development potential. They have accumulated a large amount of unlabeled data, and in terms of labeled data, they are only one order of magnitude behind Category 5. Experienced community researchers are dedicated to studying these languages. These languages have great potential to become Category 5 languages and are among those that can enjoy the superiority of digital transformation.

\textbf{Category 5 (The Winners)}:
Category 5 languages (such as Chinese and English) have developed rapidly in the NLP field and have always been in a leading position, with research time longer than languages in previous categories. Because these languages have a dominant position on the network, numerous enterprises and government institutions invest in their NLP resources and technological development. They are absolutely high-resource languages, and their users enjoy the benefits brought by the most advanced achievements and technological breakthroughs in NLP. We selected Category 5 languages and placed them in Table~\ref{tab:lang_categories}. From Table~\ref{tab:lang_categories}, we can see that 88.38\% of languages involving over 1 billion people have not enjoyed the convenience brought by NLP technological development.

It is worth noting that China is a unified multi-ethnic country with multiple ethnicities, languages, and writing systems. Besides the Han nationality, China has 55 ethnic minorities accounting for 8\% of the total population~\cite{nopss}. Among these ethnic minorities, except for the Hui nationality who mostly use Chinese language and writing, other ethnic minorities (such as Mongolian, Tibetan, Uyghur, Kazakh, Korean, etc.) use their own ethnic writing systems to varying degrees. 

The state has also decided to provide documents and simultaneous interpretation services in seven ethnic minority languages (Mongolian, Tibetan, Uyghur, Kazakh, Korean, Yi, and Zhuang) at major conferences such as the "Two Sessions," fully reflecting the ethnic policy of equality among all ethnicities. Meanwhile, CCL, organized by the Chinese Information Processing Society of China as the largest Chinese community domestically, 
encourages the publication of original research and application papers related to Chinese ethnic minority languages in computational linguistics.

Additionally, according to the requirements of international computer processing technology development, China has created encoding systems for ancient scripts such as Tibetan, Mongolian, Manchu, Dai, Kharoshthi, Tangut, 'Phags-pa, and Lisu, thus providing broader academic space for the exchange of ethnic minority ancient scripts and cultural information. China's character schemes for Jurchen script, Khitan small script, Khitan large script, Turkic script, Shui script, Dongba script, and Lisu script have received high recognition from the International Organization for Standardization, securing academic discourse power in this field.

\subsubsection{Multilingual Models}

The classification of languages also determines the research directions for different languages. For example, for Category 3 and Category 4 languages, due to the lack of sufficient labeled data, unsupervised learning methods can be used to compensate. In recent years, with the development of large models, related research mainly uses pre-training technology to learn language features from unlabeled data, achieving zero-shot or few-shot learning capabilities for target languages. Currently, there are related research results published in academia, mainly focusing on Southeast Asian languages such as Indonesian~\cite{koto2020indolem}, Vietnamese~\cite{nguyen2020phobert}, and Thai~\cite{lowphansirikul2021wangchanberta}.

Of course, relying solely on unlabeled data to improve the performance of Category 3 and Category 4 language tasks is not enough. A natural idea is to transfer language features from high-resource languages to low-resource languages, known as cross-lingual transfer learning and multilingual learning. The former refers to transfer from one source language to one target language (one-to-one), while the latter refers to transfer from multiple source languages to one target language. After the emergence of open-source large models like BLOOM/Llama, the latter has gradually become the mainstream of research. However, in the process of multilingual transfer learning, the similarity between languages greatly constrains transfer performance. Taking the transfer between English and Spanish in machine translation as an example, because the two languages are very close, with nearly 50\% overlap in words, the few-shot transfer effect and translation effect are very good. However, due to differences in tokens and other features between Chinese and English, the transfer effect is poor and can even lead to negative transfer and catastrophic forgetting problems.

Since 2023, with the development of large models, prompt engineering has become a hot research topic, roughly divided into two technical directions:

\begin{enumerate}
    \item Constructing Prompts for ICL:
    This technical direction uses prompts as input but does not fine-tune model parameters. On the multilingual side, it roughly divides into cross-lingual prompts and cross-lingual chain-of-thought prompts. Both use English as the main language for prompts but use target languages for test samples. These methods have been proven to achieve better results and have similar effects to translating test samples into English~\cite{huang2023not}.
    \item Parameter Fine-tuning Using Multilingual Prompts:
    Experimental results show that using English as prompts for fine-tuning under multilingual tasks can achieve State-of-the-Art (SOTA) results on zero-shot learning tasks in both English and non-English task sets. At the same time, using machine-translated multilingual prompts for fine-tuning can achieve better results in some languages than prompts fine-tuned with human translation~\cite{muennighoff2022crosslingual}.
\end{enumerate}

\section{Multilingual Large Language Models}
Data resources are as important in this era of large language models (LLMs) as oil and coal were in the industrial age. This section will summarize the mainstream corpora used in training MLLMs, and how to process and refine these corpora to better help researchers improve the effectiveness of MLLMs.

\subsection{Pre-training Resources}
\label{sec:dataset}
Unlike previous pre-training model technologies, large language models require massive training corpora to learn more comprehensive knowledge and content. Therefore, more and more open-source training corpora are being used to train large models. In this section, we briefly categorize the currently widely used corpus resources according to their content type into books resources, web resources, Reddit resources, Wikipedia, codes, and others~\cite{zhao2023survey}, as shown in Table~\ref{tab:language_resources}.

\begin{table}[t]
    \centering
    \scalebox{1.2}{
    \begin{tabular}{|l|l|l|l|}
        \hline
        \textbf{Resource} & \textbf{Size} & \textbf{Source} & \textbf{Time} \\ \hline
        BookCorpus & 5GB & Books & 2015-12 \\ \hline
        Gutenberg & --- & Books & 2021-12 \\ \hline
        C4 & 800GB & CommonCrawl & 2019-04 \\ \hline
        CC-Stories-R & 31GB & CommonCrawl & 2019-09 \\ \hline
        CC-NEWS & 78GB & CommonCrawl & 2019-02 \\ \hline
        REALNEWs & 120GB & CommonCrawl & 2019-04 \\ \hline
        OpenWebText & 38GB & Reddit Links & 2023-03 \\ \hline
        Pushift.io & 2TB & Reddit Links & 2023-03 \\ \hline
        Wikipedia & 21GB & Wikipedia & 2023-03 \\ \hline
        BigQuery & --- & Codes & 2023-03 \\ \hline
        The Pile & 800GB & Other & 2020-12 \\ \hline
        ROOTS & 1.6TB & Other & 2022-06 \\ \hline
    \end{tabular}}
    \caption{Summary of Main Language Resources}
    \label{tab:language_resources}
\end{table}

\begin{figure*}[t]
    \centering
    \includegraphics[width=\textwidth]{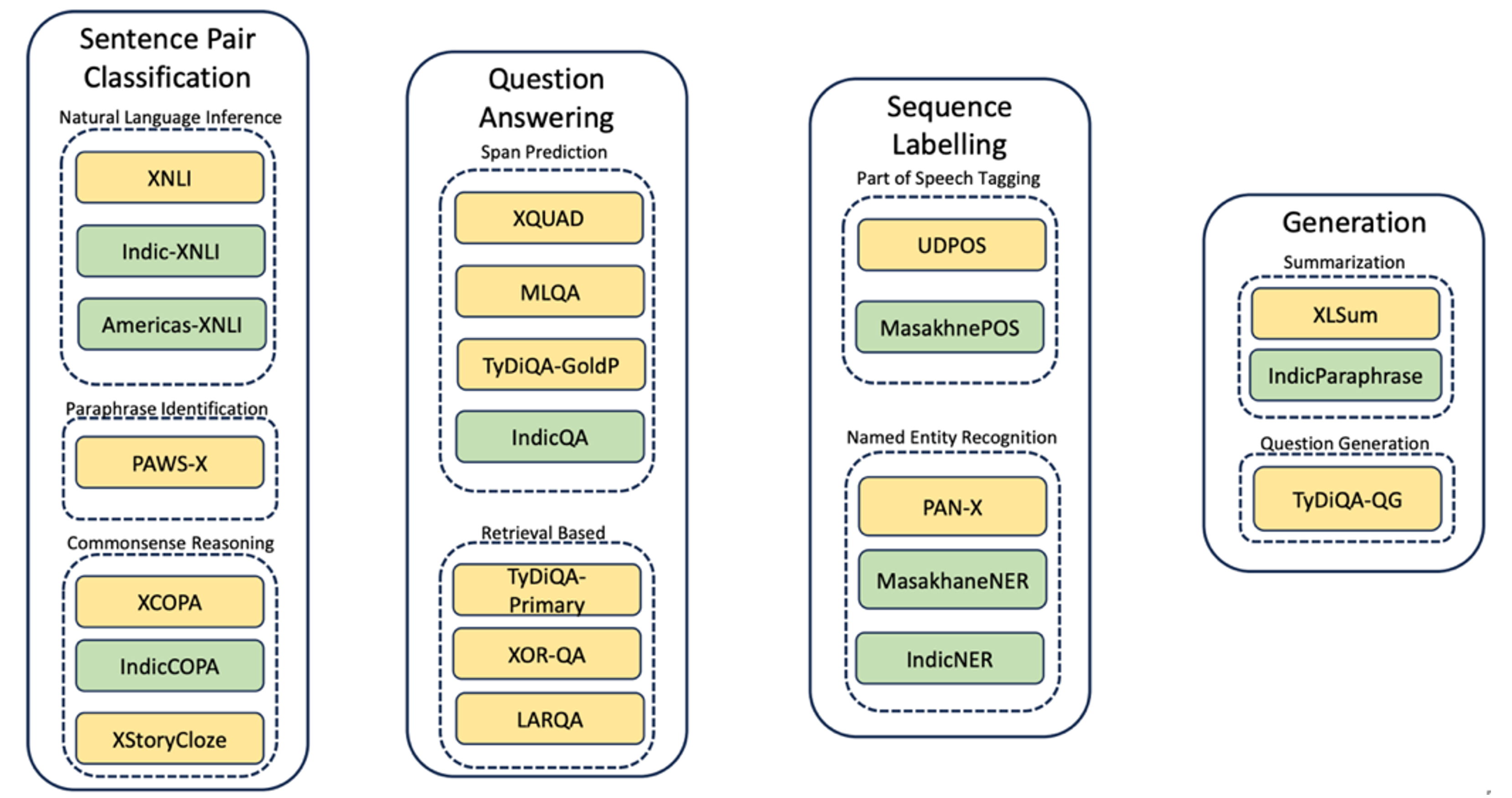}
    \caption{Evaluation Benchmarks and Related Datasets for MLLMs}
    \label{fig:mllmbenchmark}
\end{figure*}

\subsubsection{Book Resources}
BookCorpus~\cite{zhu2015aligning} is a dataset frequently used by previous pre-training models (such as GPT/GPT-2), including more than 11,000 books covering a wide range of categories (such as novels and biographies). Currently, a larger scale book resource is Project Gutenberg~\cite{project_gutenberg}, which contains over 70,000 books and is also used to train Llama and Llama2 models.

\subsubsection{Web Resources}
CommonCrawl, as one of the largest open-source web crawl databases (PB-level), is widely used in the pre-training of large language models. Due to its rich corpus resources, current language models only use subsets of data from certain time periods for training. Web crawl data contains a lot of noise and low-quality data, so it needs to be cleaned and pre-processed before use. Currently, there are several cleaned multilingual datasets available for selection: C4~\cite{raffel2020exploring}, OSCAR~\cite{abadji2022towards}, CCAligned~\cite{el-kishky2019ccaligned}.

\subsubsection{Reddit Resources}
Reddit is a social networking platform similar to an online forum where users can post and answer topics they want to discuss, and upvote or downvote answers. Those high-voted resources are valuable. OpenWebtext is a dataset formed by crawling this data. Another dataset is PushShift.io~\cite{baumgartner2020pushshift}, which is a real-time updated dataset that supports users to search, summarize, and other operations in the entire dataset, making it convenient for users to use and process the data.

\subsubsection{Wikipedia}
Wikipedia has a large number of high-quality articles in different domains and covers multiple languages. The English version of Wikipedia has been used by Llama, GPT-3, and LaMDA, while the multilingual version has been used by mBERT and XLM-100. However, because the number of documents in the multilingual version is much less than the CommonCrawl dataset, it has not been widely used by MLLMs. WikiMatrix~\cite{schwenk2019wikimatrix} is a parallel corpus extracted from Wikipedia data, containing 135 million parallel corpora, covering 1620 language pairs and 85 languages, mainly used for machine translation.

\subsubsection{Code}
For code datasets, existing work mainly focuses on crawling code with open-source licenses from the internet. Code mainly has two crawling sources: the first is public code repositories, such as GitHub; the second is code-related Q\&A platforms, such as Stack Overflow, CSDN. Google has open-sourced BigQuery~\cite{google_bigquery}, which includes a certain amount of open-source code in different programming languages. As a code large model, CodeGen~\cite{nijkamp2022codegen} used BigQuery for training, and the multilingual version of BigQuery was also used to train CodeGen-Multi.

\subsubsection{Others}
Among other corpus resources, the most distinctive are The Pile dataset and the ROOTS dataset. The Pile dataset includes over 800GB of datasets extracted from multiple data sources mentioned above, covering a subset of 22 high-quality data sources. The Pile dataset is used by various scales of large models, such as GPT-J (6B) and OpenLLaMA (7B/13B).

The ROOTS dataset is a multilingual dataset composed of many smaller datasets, covering 59 languages. The multilingual large model BLOOM used the ROOTS dataset for training. In addition, there is a Paracrawl that uses the Bitextor tool to mine parallel sentence pairs with English from web pages, mainly covering European languages and including 9 low-resource languages.

\subsubsection{Quality of Pre-training Resources}
Data quality is key to training MLLMs with excellent performance. Robust MLLMs can perform better on cross-lingual downstream tasks~\cite{liu2021crisisbert,li2023transformer}. The multilingual pre-training datasets introduced above are mostly automatically mined from the internet, and their data quality cannot be well guaranteed. Therefore, it is necessary to evaluate these corpora mined from the internet using different dimensions.

In the ACL 2023 Tutorial on MLLMs, Microsoft's researchers proposed to evaluate the quality of multilingual pre-training datasets from four aspects: multilingual distribution, data quality, source, and governance. In terms of multilingual distribution, researchers used the CommonCrawl dataset as an example, pointing out that although it includes more than 100 languages, 57 languages account for less than 0.001\%, so when using it to construct multilingual datasets, one should consider whether to generate data consistent with downstream task languages.

In terms of data quality, it is suggested that researchers constructing datasets should not only focus on low-resource corpora but also pay attention to samples that are misclassified into other languages. They may be classified into some high-resource language categories due to similarity with other languages and poor quality. At the same time, attention should be paid to machine-generated corpora and corpora containing pornographic or inappropriate information due to limited recognition tools. Researchers suggest evaluating the quality of multilingual pre-training datasets from dimensions such as data quantity, quality, number of domains, sustainability, and shareability.

\subsection{Evaluation Tasks}
The datasets introduced in Section~\ref{sec:dataset} mainly focus on unsupervised data. This section will introduce benchmarks for evaluating MLLMs on multilingual supervised datasets. The tasks in the benchmark mainly include text classification tasks (single sentence or sentence pair), QA tasks (Text-span Prediction), sequence labeling tasks, and text generation tasks, as shown in Figure~\ref{fig:mllmbenchmark}.

\begin{figure*}[t]
    \centering
    \includegraphics[width=\textwidth]{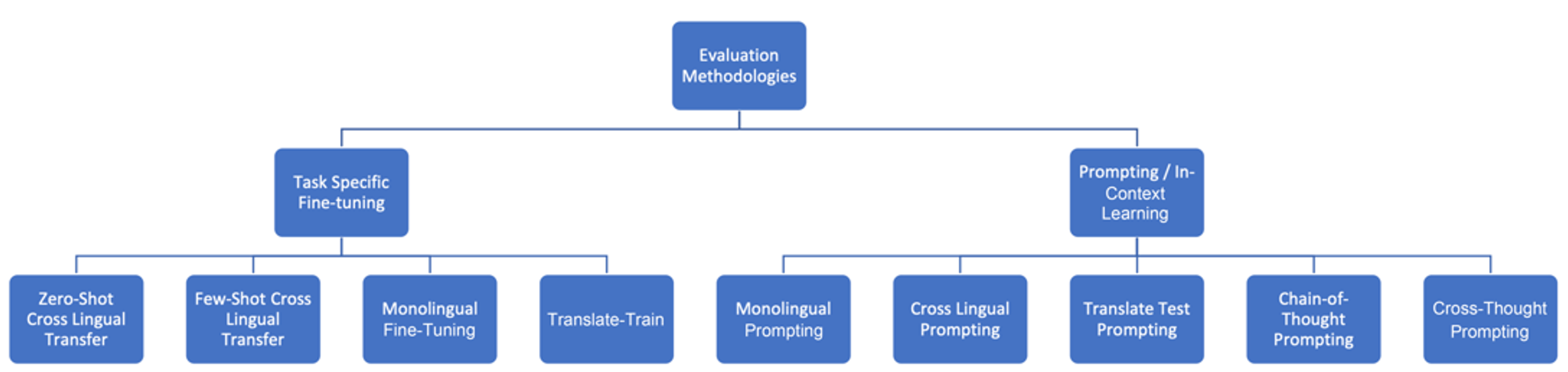}
    \caption{Evaluation methods for MLLMs}
    \label{fig:mllmeval}
\end{figure*}

\subsubsection{Datasets}
Constructing robust and comprehensive evaluation tasks can help us better understand the effectiveness of large models. This type of evaluation is a very active research area in English, such as the GLUE and the more difficult SuperGLUE evaluation dataset benchmark, as well as the recently established multilingual evaluation datasets XTREME, XTREME-R, and XGLUE based on them. The goal of multilingual evaluation datasets is to cover more diverse tasks and languages so that we can better evaluate the generalization performance of MLLMs. At the same time, some researchers are also dedicated to building evaluation datasets for specific languages, such as IndicXTREME for Indian languages.

In general, the construction of multilingual evaluation datasets lacks linguistic diversity and rarely covers low-resource languages. Before the emergence of large models, English evaluation datasets were usually translated into multiple target languages through machine translation to create evaluation datasets. This method cannot generate natural, representative target languages, which affects the validity of the evaluation. 

To better measure the performance of MLLMs on cross-lingual zero-shot and few-shot tasks, multiple research institutions collaborated to build the xP3~\cite{muennighoff2022crosslingual} dataset, covering 46 languages, including English and machine-translated prompts; Microsoft's team built the MEGA~\cite{ahuja2023mega} evaluation dataset, including 16 evaluation tasks, covering more than 70 languages; the University of Washington, Google, and Allen AI jointly released their multilingual evaluation dataset BUFFET~\cite{asai2023buffet}, including 15 evaluation tasks, covering 54 languages, and additionally providing fixed few-shot sets and instructions to better measure the effectiveness of MLLMs on few-shot cross-lingual transfer tasks.

In addition researchers also released multilingual dialogue datasets for task-oriented dialogues, such as GlobalWoZ~\cite{ding2021globalwoz} and X-RiSAWOZ~\cite{moradshahi2023xrisa}, to help developers better develop and evaluate multilingual task-oriented dialogue systems.

\subsubsection{Methods}
Traditional model evaluation methods (Evaluation Methodologies) use supervised training corpora to fine-tune pre-trained models, with the pipeline being pre-training + fine-tuning. MLLMs have inherent zero-shot and few-shot learning capabilities, and their evaluation methods can be divided into two categories: one uses the original evaluation method for task-related fine-tuning, and the other is based on prompt in-context learning (ICL), with the pipeline changing to pre-training + prompting + prediction. Task-related fine-tuning requires using training corpora to update model parameters, while ICL does not need to update model parameters, only requiring the design of different prompts for the model to return corresponding results. The evaluation methods for MLLMs are shown in Figure~\ref{fig:mllmeval}.

\textbf{Task-Specific Fine-tuning:}
\begin{itemize}
    \item Zero-Shot Cross Lingual Transfer: First perform task-related fine-tuning on one language, then evaluate using the test set of another language.
    \item Few-Shot Cross Lingual Transfer: First fine-tune parameters on English and a small amount of target language, then evaluate on the target language test set.
    \item Monolingual Fine-tuning: Only fine-tune parameters on the full target language.
    \item Translate-Train: Use machine-translated target language for parameter fine-tuning.
\end{itemize}

\textbf{Prompt-based In-Context Learning:}

This evaluation method hopes to construct appropriate prompts to activate the capabilities of large models to help people solve current tasks. It mainly consists of different dimensions: input prompts, task-related few-shot examples generalized through templates, and answers. Table~\ref{tab:prompt-icl-tasks} shows prompts and few-shot data templates for different tasks.


\begin{table*}[t]
\begin{center}
\scalebox{1.15}{
\begin{tabular}{|c|c|l|l|c|}
\hline
\textbf{Type} & \textbf{Task Example} & \multicolumn{1}{c|}{\textbf{Input ({[}X{]})}} & \multicolumn{1}{c|}{\textbf{Template}} & \textbf{Answer ({[}Z{]})} \\ \hline
\multirow{3}{*}{\begin{tabular}[c]{@{}c@{}}Text \\ Classification\end{tabular}} & Sentiment & I love this movie. & {[}X{]} The movie is {[}Z{]}. & \begin{tabular}[c]{@{}c@{}}great, fantastic, ...\end{tabular} \\ \cline{2-5} 
 & Topics & He prompted the LM. & {[}X{]} The text is about {[}Z{]}. & \begin{tabular}[c]{@{}c@{}}sports, science, ...\end{tabular} \\ \cline{2-5} 
 & Intention & What is taxi fare to Denver? & {[}X{]} The question is about {[}Z{]}. & \begin{tabular}[c]{@{}c@{}}quantity, city, ...\end{tabular} \\ \hline
\begin{tabular}[c]{@{}c@{}}Text-span \\ Classification\end{tabular} & \begin{tabular}[c]{@{}c@{}}Aspect \\ Sentiment\end{tabular} & Poor service but good food. & {[}X{]} What about service? {[}Z{]}. & \begin{tabular}[c]{@{}c@{}}Bad, Terrible, ...\end{tabular} \\ \hline
\begin{tabular}[c]{@{}c@{}}Text-pair \\ Classification\end{tabular} & \begin{tabular}[c]{@{}c@{}}Natural Language \\ Inference\end{tabular} & \begin{tabular}[c]{@{}l@{}}{[}X1{]}: An old man with ... \\ {[}X2{]}: A man walks ...\end{tabular} & {[}X1{]} ? {[}Z{]}, {[}X2{]} & \begin{tabular}[c]{@{}c@{}}Yes, No, ...\end{tabular} \\ \hline
Tagging & \begin{tabular}[c]{@{}c@{}}Named Entity \\ Recognition\end{tabular} & \begin{tabular}[c]{@{}l@{}}{[}X1{]}: Mike went to Paris. \\ {[}X2{]}: Paris\end{tabular} & {[}X1{]}{[}X2{]} is a {[}Z{]} entity. & \begin{tabular}[c]{@{}c@{}}organization, location, ...\end{tabular} \\ \hline
\multirow{2}{*}{\begin{tabular}[c]{@{}c@{}}Text\\ Generation\end{tabular}} & Summarization & Las Vegas police ... & {[}X{]} TL;DR: {[}Z{]} & \begin{tabular}[c]{@{}c@{}}The victim ...\\A woman ...\end{tabular} \\ \cline{2-5} 
 & Translation & Je vous aime. & French: {[}X{]} English: {[}Z{]} & \begin{tabular}[c]{@{}c@{}}I love you., I fancy you.,\\...\end{tabular} \\ \hline
Regression & \begin{tabular}[c]{@{}c@{}}Textual \\ Similarity\end{tabular} & \begin{tabular}[c]{@{}l@{}}{[}X1{]}: A man is smoking. \\ {[}X2{]}: A man is skating.\end{tabular} & {[}X1{]} {[}Z{]}, {[}X2{]} & \begin{tabular}[c]{@{}c@{}}Yes, No, ...\end{tabular} \\ \hline
\end{tabular}}
\end{center}
\caption{Prompts and few-shot data templates for different tasks}
\label{tab:prompt-icl-tasks}
\end{table*}

\begin{figure*}[t]
    \centering
    \includegraphics[width=.8\textwidth]{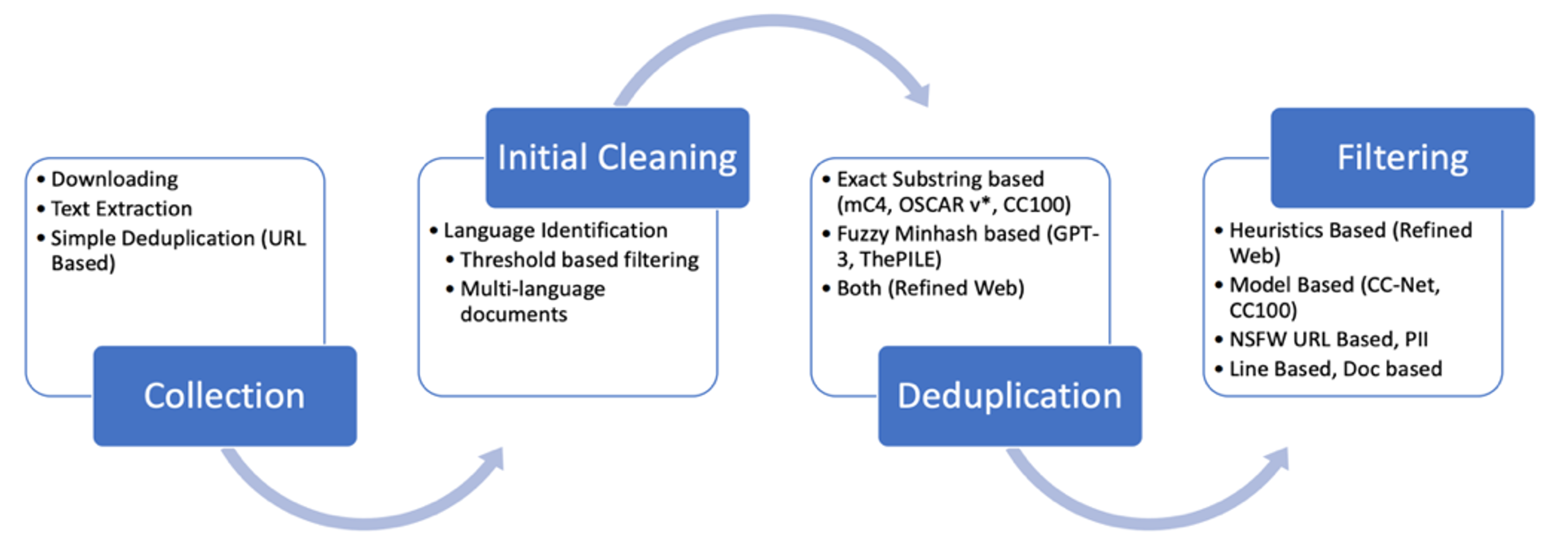}
    \caption{Pre-training data processing flow for MLLMs}
    \label{fig:mllmdataprocessing}
\end{figure*}

\begin{itemize}
    \item Monolingual Prompting: Prompts, few-shot inputs, test data, and answers are all in the target language.
    \item Cross Lingual Prompting: Prompts and few-shot inputs are in the source language, test data and answers are in the target language.
    \item Chain-of-Thought Prompting: For problems that focus on multi-step logical reasoning, adding "Step-by-Step" in the prompt helps improve the accuracy of large model answers. In this type of prompting technique, prompts, few-shot inputs, test data, and responses are all in the target language.
    \item Cross-Thought Prompting: This technique takes advantage of large models' proficiency in English. The input uses the target language, first letting the multilingual large model rephrase the question in English in the prompt, then using English for CoT reasoning, and the final response is also in English. Experiments have shown that this method performs better in cross-lingual reasoning than using only English input or translating the target language into English.
\end{itemize}

\section{Optimizing MLLMs}
After introducing the pre-training data and evaluation tasks for MLLMs, this section will discuss how to pre-process the pre-training data and tokenize it, as well as the structure and effectiveness of various MLLMs.

\subsection{Data Pre-processing Flow}
The pre-processing of pre-training data for MLLMs is not significantly different from that for monolingual large models. As shown in Figure~\ref{fig:mllmdataprocessing}, they all go through steps such as data collection, initial data cleaning (language detection), document de-duplication, and quality filtering (privacy filtering). The only difference is that during the initial data cleaning process, appropriate thresholds need to be set to identify the languages we need and retain a certain proportion of multilingual documents for model training.

\begin{figure}[t]
    \centering
    \begin{subfigure}[b]{0.45\textwidth}
        \includegraphics[width=\textwidth]{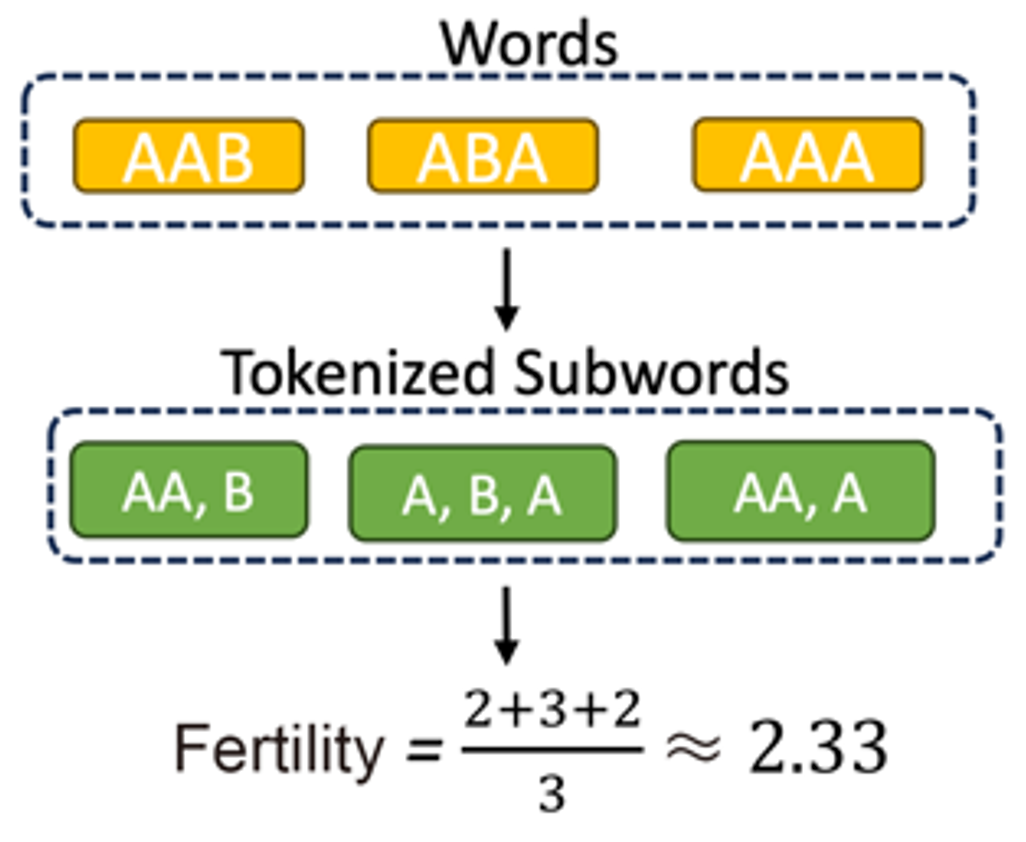}
        \caption{Subword Fertility Calculation}
        \label{fig:fertilitysubfig1}
    \end{subfigure}
    \hfill
    \begin{subfigure}[b]{0.45\textwidth}
        \includegraphics[width=\textwidth]{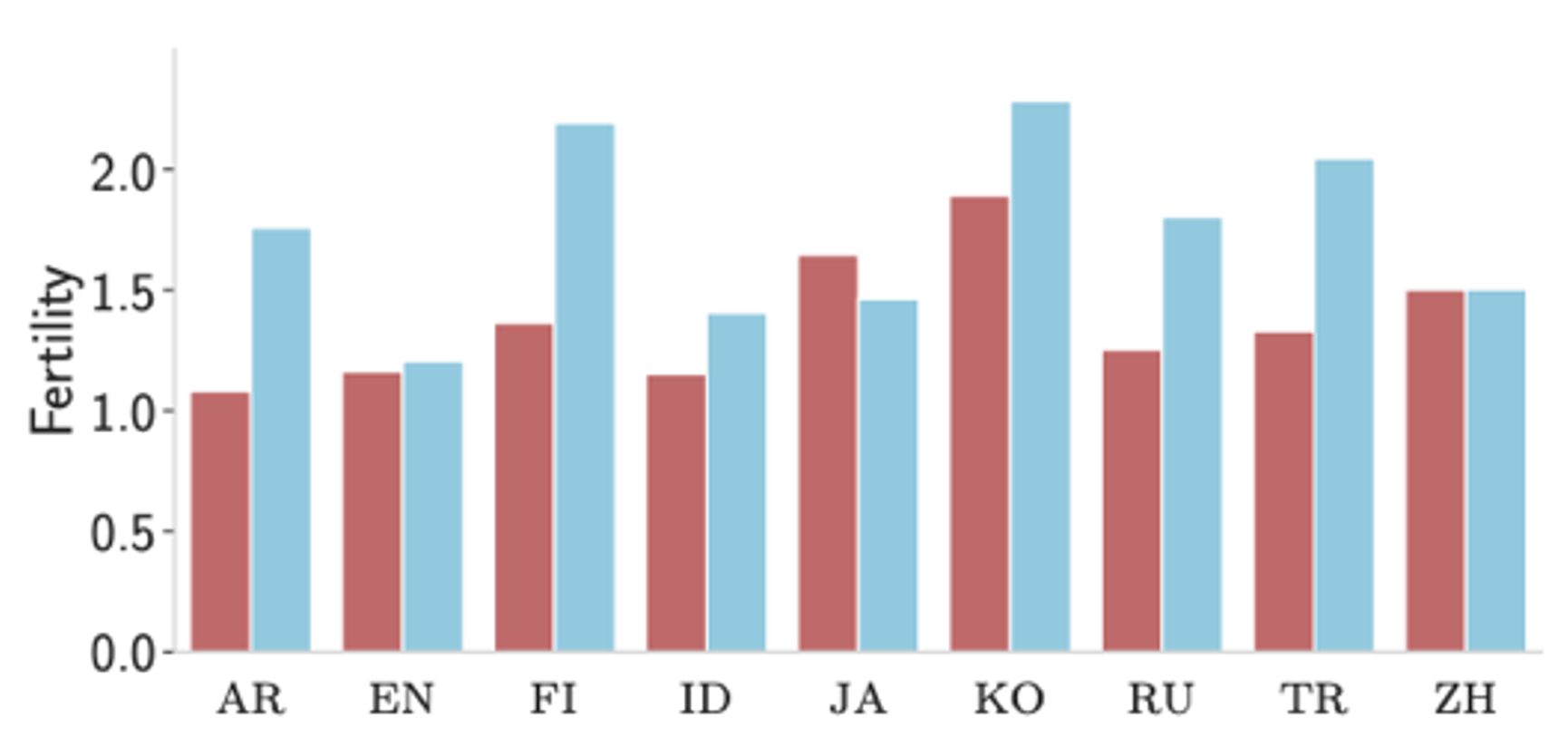}
        \caption{Cross-linguistic Comparison of Subword Fertility Rates}
        \label{fig:fertilitysubfig2}
    \end{subfigure}
    \caption{Subword Fertility Analysis: Definition and Cross-linguistic Comparison
    (a) Example calculation of subword fertility ratio  (b) Subword fertility rates for different languages including Arabic (AR), English (EN), Finnish (FI), Indonesian (ID), Japanese (JA), Korean (KO), Russian (RU), Turkish (TR), and Chinese (ZH)}
    \label{fig:fertility}
    \vspace{-.4cm}
\end{figure}

\begin{figure}[t]
    \centering
    \begin{subfigure}[b]{0.45\textwidth}
        \includegraphics[width=\textwidth]{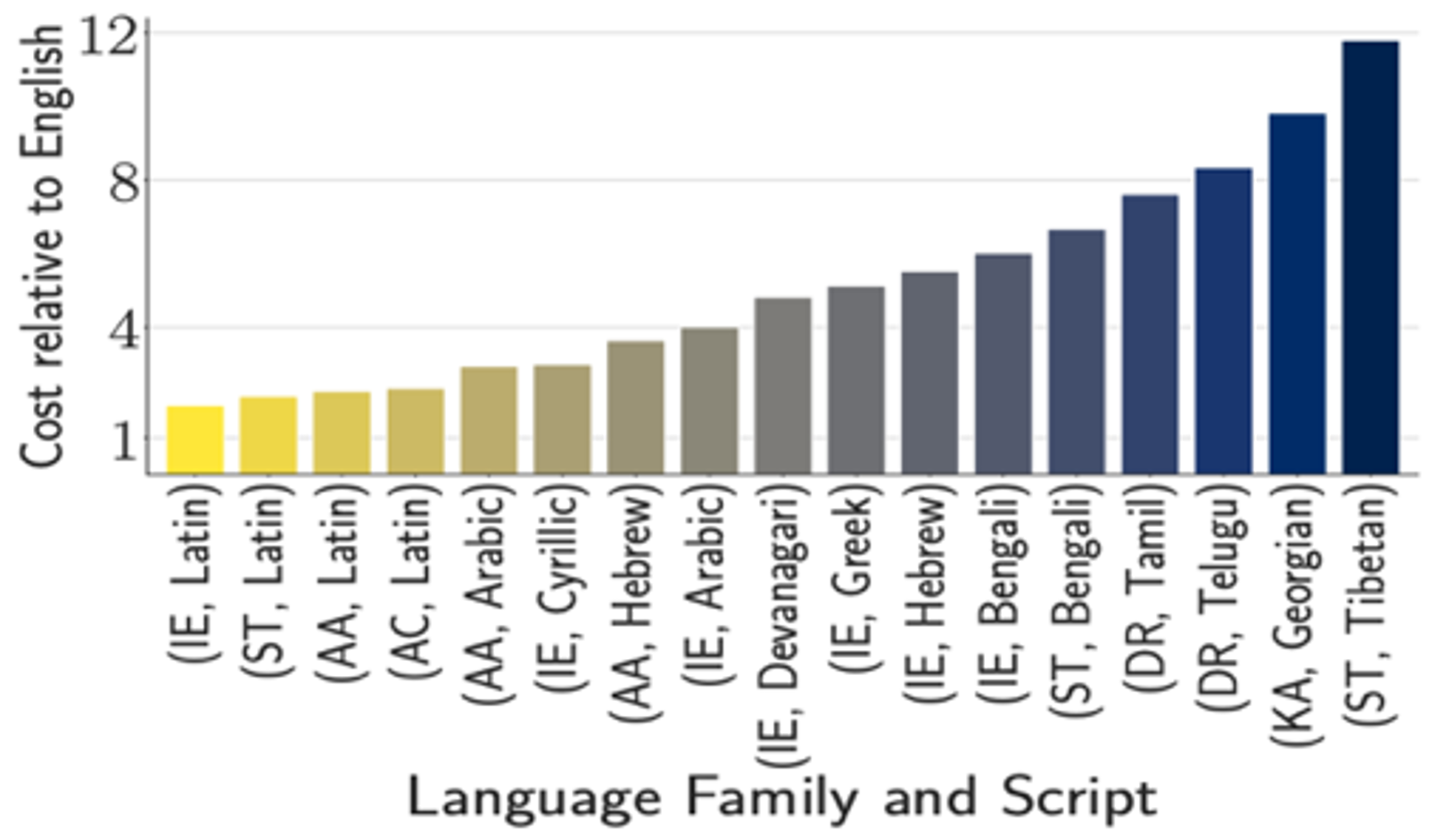}
        \caption{Relative Inference Cost Across Language Families and Scripts Compared to English}
        \label{fig:langsub1}
    \end{subfigure}
    \hfill
    \begin{subfigure}[b]{0.45\textwidth}
        \includegraphics[width=\textwidth]{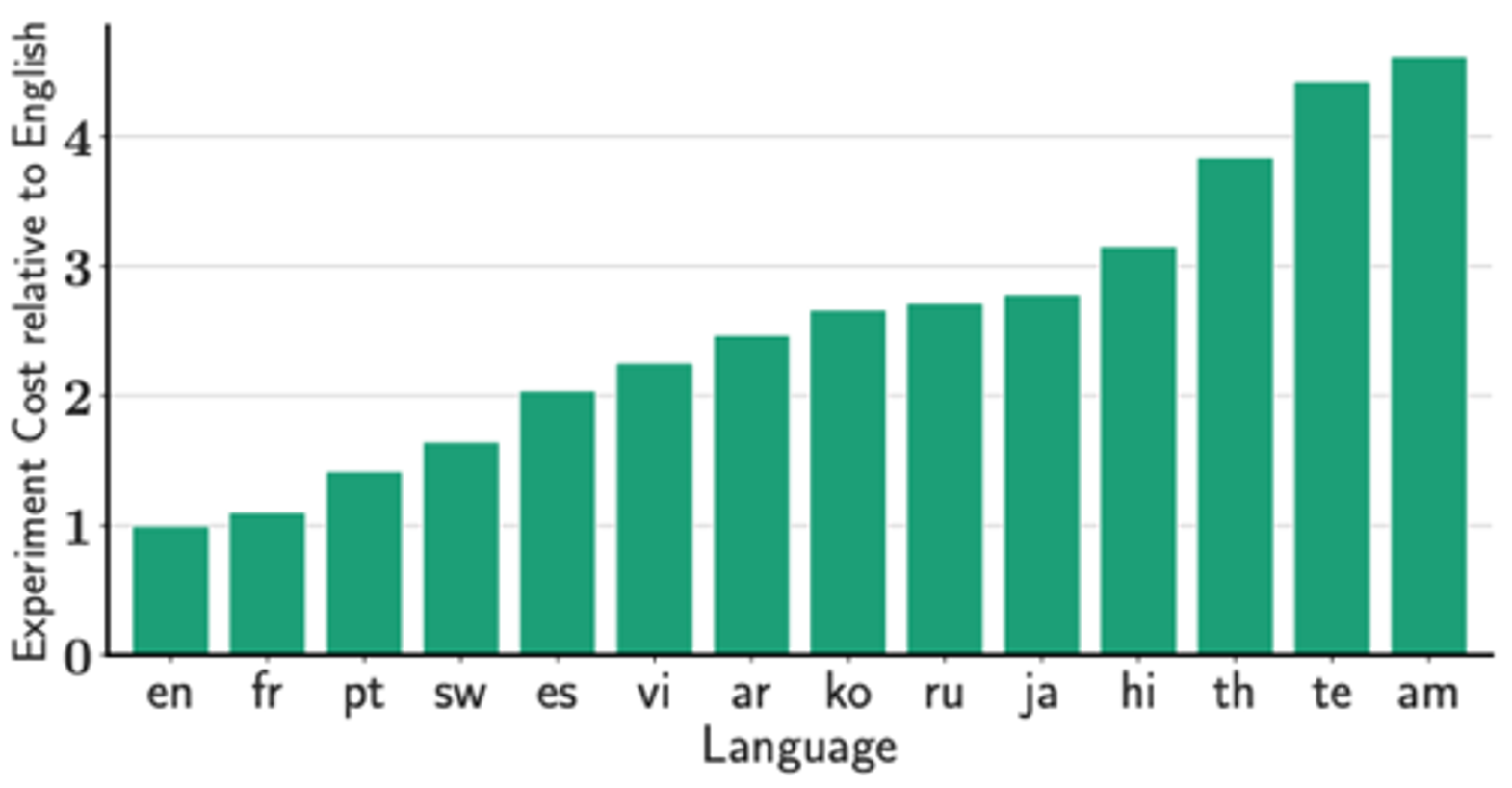}
        \caption{Experimental Cost Comparison Across Different Languages Relative to English}
        \label{fig:langsub2}
    \end{subfigure}
    \caption{Relationship Between Language Model Inference Cost and Multilingual Tokenizer}
    \label{fig:lang}
    \vspace{-.4cm}
\end{figure}

\begin{figure*}[t]
    \centering
    \includegraphics[width=\textwidth]{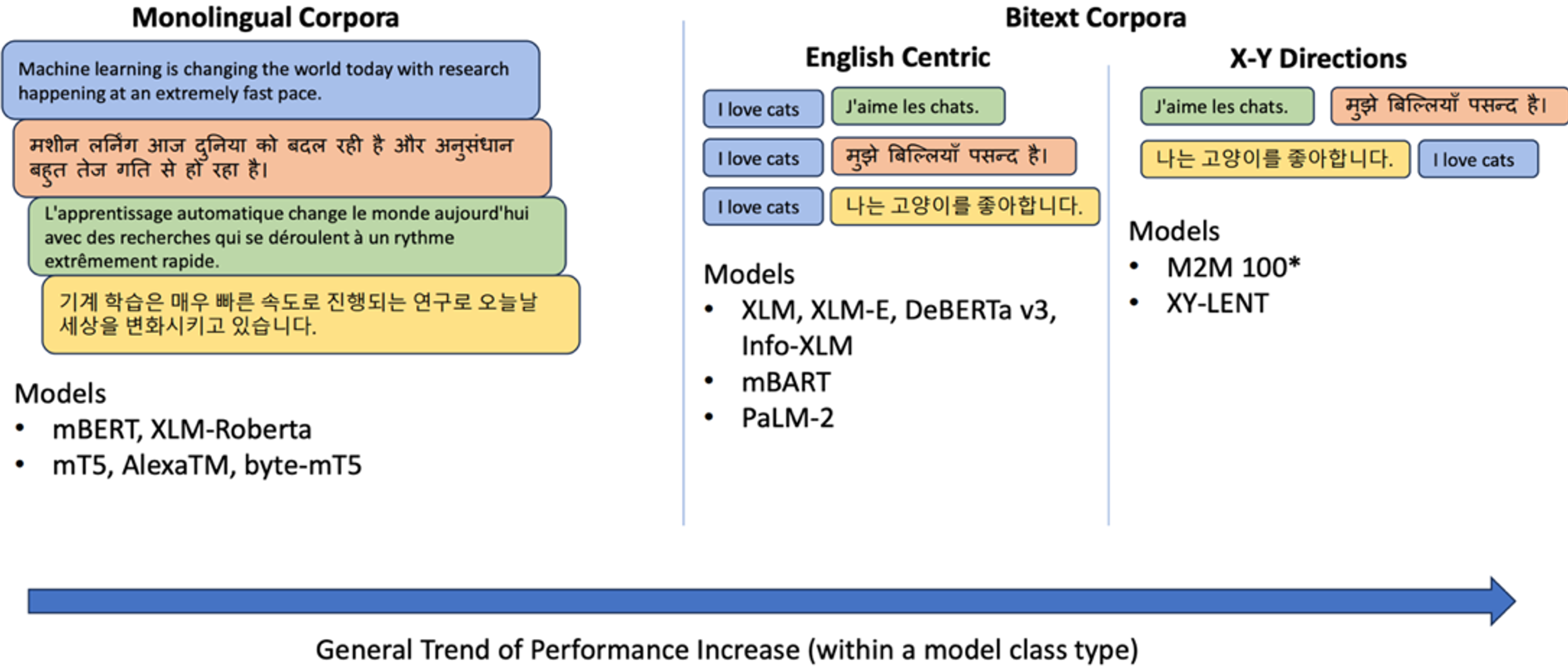}
    \caption{Comparison of English-centric parallel sentence training data formats and models}
    \label{fig:parallel-training}
\end{figure*}

\subsection{Tokenizer}
After the corpus has gone through data pre-processing, it needs to be processed by a tokenizer before it can be converted into a data format directly usable by MLLMs. Below, we will discuss commonly used tokenizers and their potential issues in multilingual scenarios.

\subsubsection{Tokenizer Selection}
There are mainly two types of mainstream tokenizers, one is subword-based tokenizers, of which the following two are representative:

\textbf{BPE algorithm~\cite{huggingface2023nlp}:} learining learns how to merge the two most common and consecutive tokens in the training dictionary into a new token and add it to the dictionary until the dictionary capacity is met. This method is also used by GPT, BLOOM, and LLaMA series.

\textbf{SentencePiece~\cite{sentencepiece2023}}: an open-source code library that implements tokenization algorithms, such as BPE and Unigram Tokenization, and also non-subword tokenizers, such as character-based and letter-based tokenizers. Large models such as XGLM, mT5, and PaLM series all use this library as a tokenizer.

The other type is byte-based tokenizers, BBPE~\cite{wang2020neural}. When our dictionary capacity is limited, tokenizing with bytes as the smallest token can encode all possible sequence information and has good fault tolerance for spelling. Research has trained a multilingual large model mT5 without subwords using UTF-8 encoding, showing good results on multilingual test sets. This method can be used to train large language models without subwords (Latin-based languages) because UTF-8 encoding of Latin-based languages will be smaller than encoding Chinese, Japanese, and Korean. However, this method requires a deeper encoder and needs the model to accept longer context information. Llama2 uses this method for non-UTF-8 characters.

\subsubsection{Potential Issues with Tokenizers}
Compared to monolingual LMs, MLLMs pay more attention to the quality of tokenizers because MLLMs need to use tokens to represent more language sequences with a limited vocabulary. Ahia et. al.~\cite{ahia2023tokenization} has conducted quantitative analysis from the perspective of subword fertility. The definition and calculation method of subword fertility are shown in Figure~\ref{fig:fertilitysubfig1}. Fertility is mainly used to calculate the average length of a natural word being split into subwords, with the minimum fertility being 1, indicating that each natural word is a subword. As can be seen from Figure~\ref{fig:fertilitysubfig2}, the fertility of the multilingual large model mBERT in English (EN), Chinese (ZH), Indonesian (ID), and Japanese (JA) is basically the same as that of monolingual large models, but it is significantly higher than that of their respective monolingual large models in Korean (KO), Arabic (AR), Finnish (FI), Russian (RU), and Turkish (TR). This is because mBERT's training corpus contains a large amount of high-resource corpora such as English, and English is a morphologically poor language compared to Arabic, Finnish, Russian, and Turkish. When the tokenizer uses the same vocabulary across multiple languages, it leads to the tokenizer not being able to learn the optimal word combination in the corresponding low-resource corpora.

As the subword fertility for certain languages increases, we have to consider the inference cost of MLLMs. Research has shown that using prompt-based ICL to call MLLMs for prediction in low-resource languages will greatly increase the cost of use because we will decompose the sequence into more tokens, increasing the number of tokens in the context and output. As shown in Figure~\ref{fig:lang}, if English or Latin-based languages are used as the benchmark, it can be seen from Figure~\ref{fig:langsub1} that the calling cost of large models for Latin-based languages is the smallest, and from Figure~\ref{fig:langsub1} that the calling cost for other languages is generally higher than English. This is because the tokenizer has not been well-trained on the corresponding languages.

In addition to the above quantifiable evaluations, tokenizers also have problems for some languages that need to be tokenized, such as Chinese, Japanese, and Thai~\cite{fujii2023tokenizers,chung2020improving}.

\begin{figure*}[t]
    \centering
    \includegraphics[width=\textwidth]{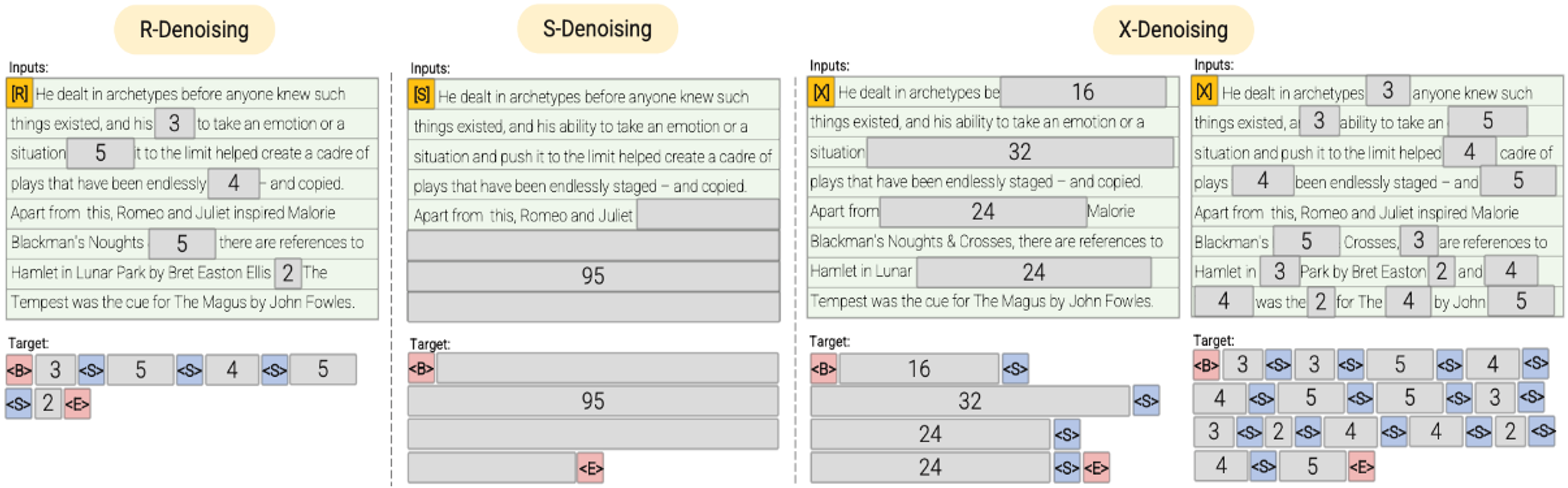}
    \caption{Training objectives of different language models}
    \label{fig:training-obj}
\end{figure*}

\subsection{Training Data Format and Sampling Strategy}
After the corpus has gone through tokenizer processing, before it can be fed into MLLMs, the models need to consider how to organize multilingual samples and the sampling strategy for different language samples. This section will discuss monolingual samples, English-centric parallel sentence pairs, and multilingual parallel sentence pairs. In general, using multilingual parallel sentence pairs for training will yield better results compared to using monolingual samples or English-centric parallel sentence pair samples. The reason for sampling languages is to better balance the impact of model scale and language scale on the effectiveness of MLLMs, avoiding the curse of multilinguality phenomenon mentioned in Section 1.

\subsubsection{Monolingual Samples}
Monolingual sampling refers to the strategy of treating the training data from each language as an independent pre-training task sample during the training process. Different sampling strategies are employed to sample from different languages in the corpus to form a batch for alternative training. This is the training data format used by models like GPT-3/LLaMA series/BLOOM. The reason for sampling across languages is that there are three multilingual sampling strategies.

\textbf{The first approach} imposes no restrictions and samples directly based on the empirical distribution $p_l$ formed by the number of samples for each language, where $p_l$ represents the sample quantity for the $l$-th language.

\begin{equation}
    p_l = \frac{n_l}{\sum_{l'\in L} n_{l'}}
    \tag{7-1}
    \end{equation}

\textbf{The second approach} uses a temperature-based sampling strategy~\cite{xue2020mt5,conneau2019unsupervised}, generally applied to models with fewer than 10 billion parameters. This sampling strategy first uses a sampling function $q_l$ and a temperature parameter $\tau$ to normalize the original $p_l$ distribution. By setting different temperatures, it ensures that low-resource languages can be oversampled and high-resource languages can be undersampled, thus balancing the proportion between different languages in the training data.

\begin{equation}
    q_l = \frac{p_l^{1/\tau}}{\sum_{l'\in L} p_{l'}^{1/\tau}}
    \tag{7.2}
    \end{equation}

\textbf{The third approach} is the Unimax method proposed by ICLR in 2023~\cite{chung2023unimax}, which aims to distribute the training budget more evenly based on the number of tokens in each language's data, thus eliminating the need for manually determining the value of $\tau$ and introducing randomness to the experiment. This method first allocates the budget starting from low-resource languages to form batches for model training. Experiments have shown that this method performs better than using temperature-based sampling functions.

\subsubsection{English-centric Parallel Sentence Pairs}
The training data format of English-centric parallel sentence pairs and the corresponding model are shown in Figure~\ref{fig:parallel-training}. Using English as the intermediate language, it samples the target language corpus aligned with that sample to form parallel sentence pairs and inputs them into the model for Masked LM task pre-training. This method borrows the idea of machine translation, hoping to let the pre-training model learn the connection between different languages and English by masking token information of different languages.

\subsubsection{Multilingual Parallel Sentence Pairs}
\label{sec:parallel}
Compared with the English-centric parallel sentence pair method, multilingual parallel sentence pairs no longer use English as the intermediate language to ensure that the model can learn direct relationships between different languages during the training process. This is because the morphologically poor feature of English determines that it may not necessarily be a good bridge language~\cite{paul2013pivot}. In October 2022, the T-ULRv6~\cite{bing2022turing} model trained using this method achieved SOTA levels on the previously mentioned multilingual benchmark XTREME and monolingual GLUE, proving that MLLMs can achieve the best results on both English and multilingual evaluation tasks simultaneously.

\subsection{Training MLLMs}
This section will introduce the model training part, divided into model pre-training tasks and fine-tuning tasks. For the pre-training tasks of MLLMs, we mainly focus on two aspects: the training objectives and the structure of MLLMs.

\subsubsection{Constructing Objectives}
Before introducing the structure of MLLMs, we briefly introduce three methods for constructing training objectives. The training objectives of different language models are shown in Figure~\ref{fig:training-obj}. Different training objectives form prediction targets to calculate loss by covering different parts of the input, and then use it to update the model parameters.

 \textbf{Regular-Denoising (R-Denoising)}: Generally used for MLM masking strategy, the training objective is usually to first replace tokens within a certain range with independent masks, and then train the model to predict them, such as BERT.

 \textbf{Specific-Denoising (S-Denoising)}: This method divides a given sentence into prefix and suffix parts, with the prefix part as context and the suffix part as the training target, most commonly used for text generation and machine translation tasks.

\textbf{Specific-Denoising (S-Denoising)}: This method~\cite{tay2022ul2} can be viewed as an extension of Regular-Denoising, where the masking window is changed from tokens to sentences, with the final training objective being to predict each covered sentence. This approach is commonly adopted for multi-turn dialogue tasks when training MLLMs. In practical applications, the Llama2 model only used method (2), while Llama2-chat~\cite{touvron2023llama2} employed both methods (2) and (3). Google's PaLM 2 model utilized all three methods (1)--(3), determining different training objectives by adding R-S-E flags before each input~\cite{anil2023palm2}.

\subsubsection{Structure of MLLMs}
Transformer~\cite{vaswani2017attention} is currently the most commonly used multilingual large model structure, mainly composed of encoder and decoder structures. From their combination, they can be divided into encoder, encoder-decoder, and decoder. Currently, MLLMs mainly use decoder, encoder-decoder, and prefix language model (Prefix LM) structures. The early pre-training model BERT used the encoder structure, which is now rarely used independently, so this article mainly introduces these three structures of MLLMs.

\begin{figure*}[ht]
    \centering
    \begin{subfigure}[c]{0.33\textwidth}
        \includegraphics[width=\textwidth]{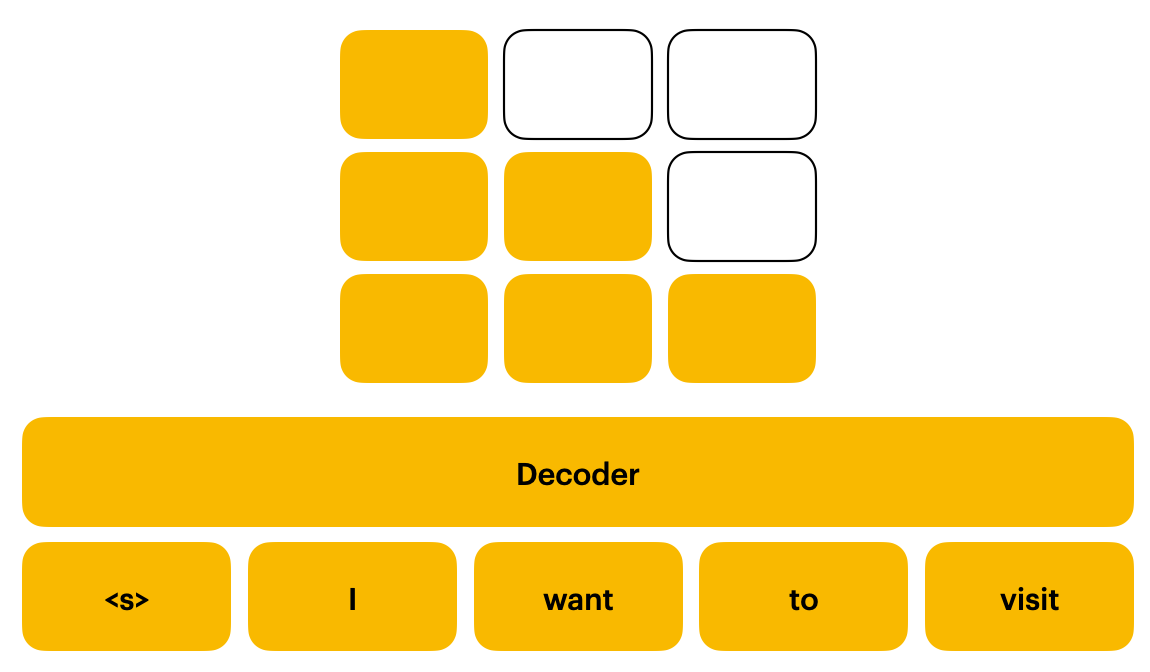}
        \caption{Decoder-only}
        \label{fig:lmsub1}
    \end{subfigure}
    \vrule width .2pt height 2cm
    \hfill
    \begin{subfigure}[c]{0.3\textwidth}
        \includegraphics[width=\textwidth]{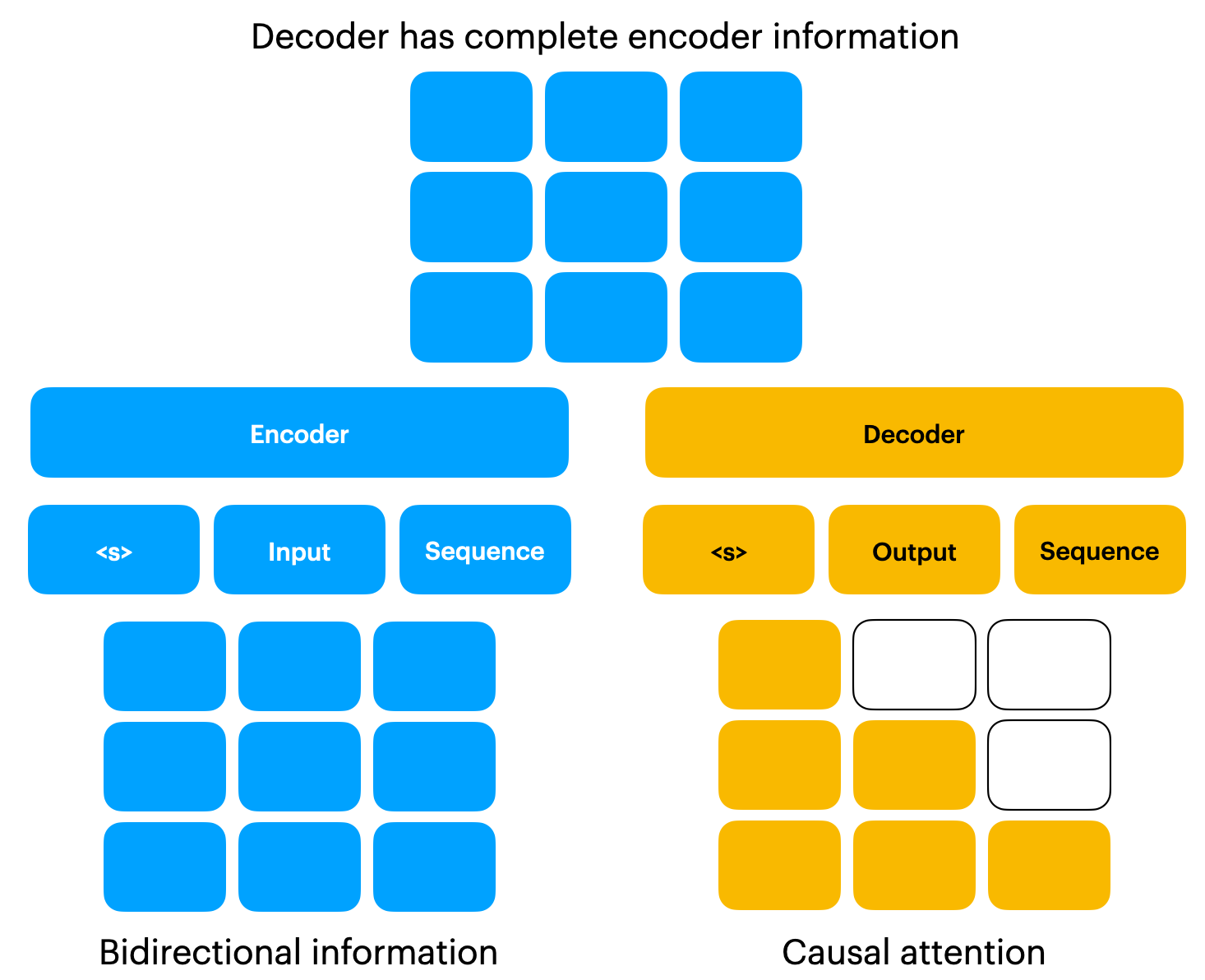}
        \caption{Encoder-Decoder}
        \label{fig:lmsub2}
    \end{subfigure}
    \vrule width .2pt height 2cm
    \hfill
    \begin{subfigure}[c]{0.32\textwidth}
        \includegraphics[width=\textwidth]{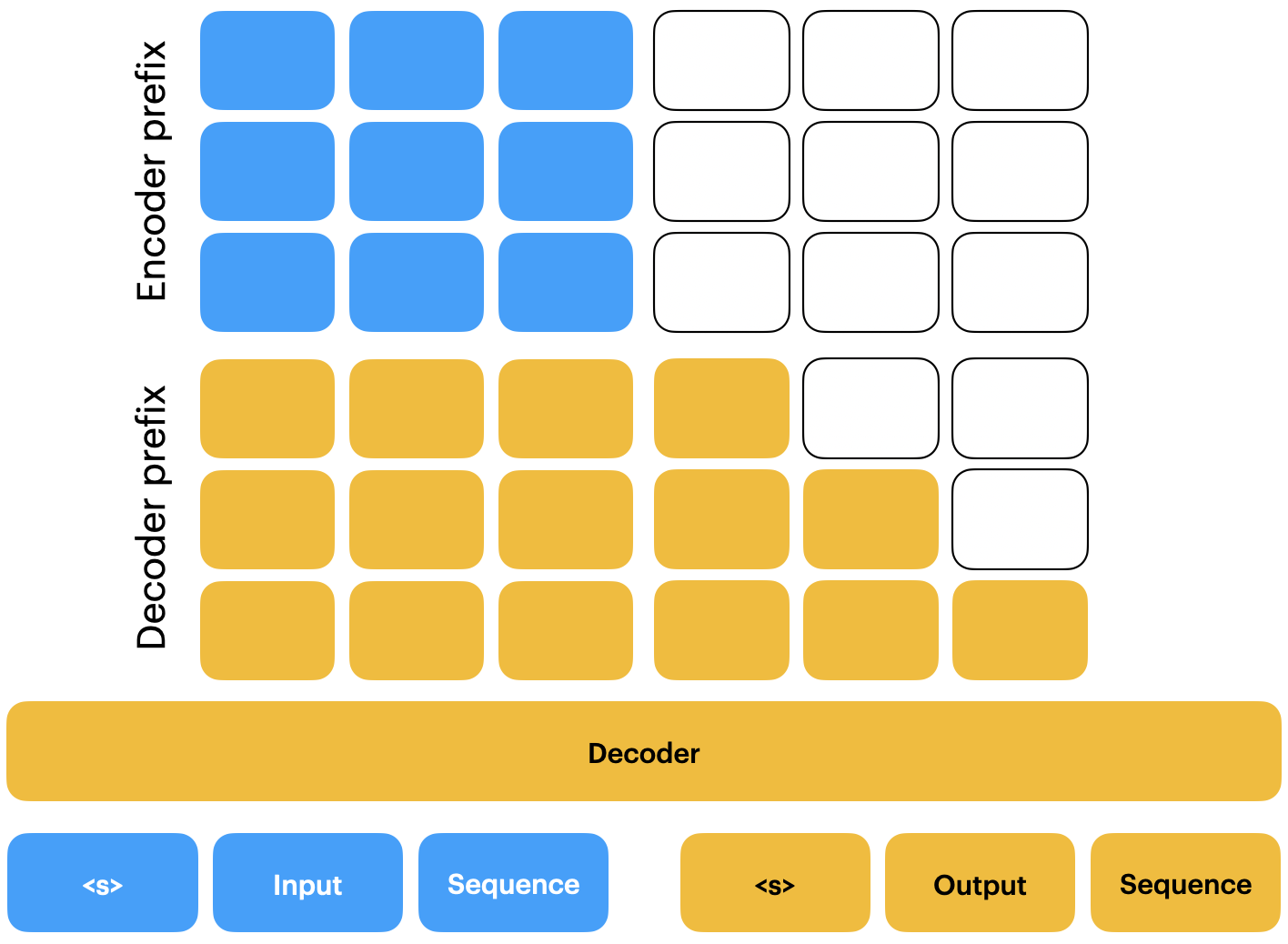}
        \caption{Prefix language model}
        \label{fig:lmsub3}
    \end{subfigure}
    \caption{Different Structures of Language Models}
    \label{fig:lm}
\end{figure*}

\textbf{Decoder Structure:}
The decoder structure is shown in Figure~\ref{fig:lmsub1}. This structure is currently employed by mainstream MLLMs, allowing the model to be trained like traditional autoregressive language models using unidirectional attention mechanisms for next token prediction (NTP). This architecture is utilized by Llama2, GPT series, BLOOM, and XGLM models. Recent research suggests that model performance can be enhanced through decoder-less training~\cite{tay2022transcending} or instruction fine-tuning~\cite{chung2022scaling} on decoder-based models. However, these methods have primarily been tested on English-centric models, and their effectiveness on MLLMs remains unclear. The PaLM 2 model has reported excellent results using similar approaches for multilingual model tasks, suggesting this could be a viable optimization strategy for MLLMs.

\textbf{Encoder-Decoder Structure:}
The traditional encoder-decoder structure is shown in Figure~\ref{fig:lmsub2}. This structure uses the encoder and decoder of the Transformer. It first inputs the token sequence into the encoder to obtain a sequence vector of the same length as the input, and then uses it as the input of the decoder. The decoder uses cross-attention mechanisms, using bidirectional attention mechanisms for input to pay attention to all contexts of the input. Bidirectional attention mechanism is an efficient strategy for utilizing data because it can use information before and after this token when predicting tokens. However, this method is better at natural language understanding tasks rather than the natural language generation tasks that large models do now, so it is less used independently in large models. For the output sequence, it uses unidirectional attention mechanisms to prevent the model from paying attention to information after the predicted token. mT5 continues to use this structure.

\begin{figure}[t]
    \centering
    \includegraphics[width=.45\textwidth]{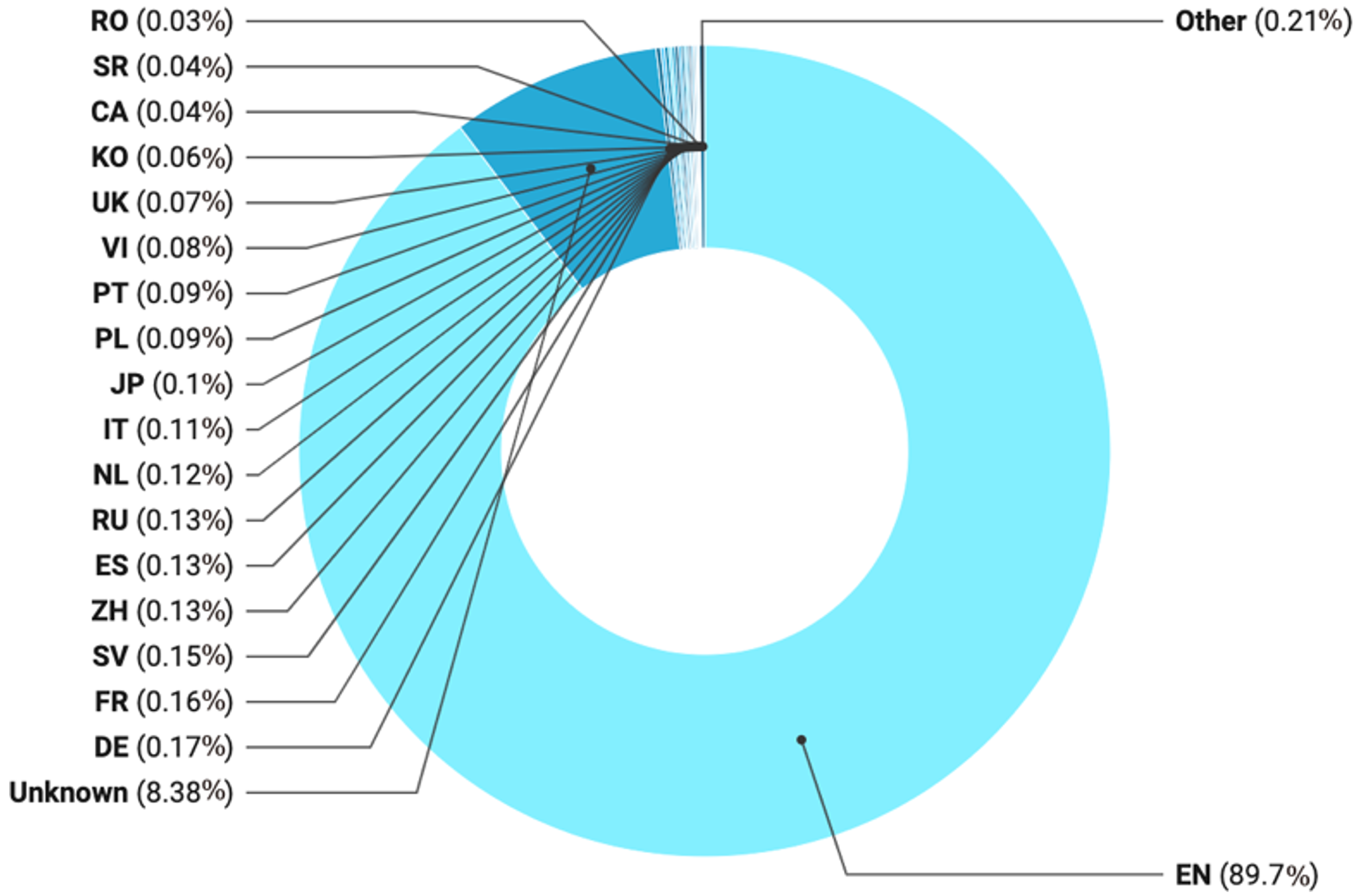}
    \caption{Proportion of each language in Llama2's training corpus}
    \label{fig:llama2-dist}
    \vspace{-.5cm}
\end{figure}

\begin{figure}[t]
    \centering
    \includegraphics[width=.48\textwidth]{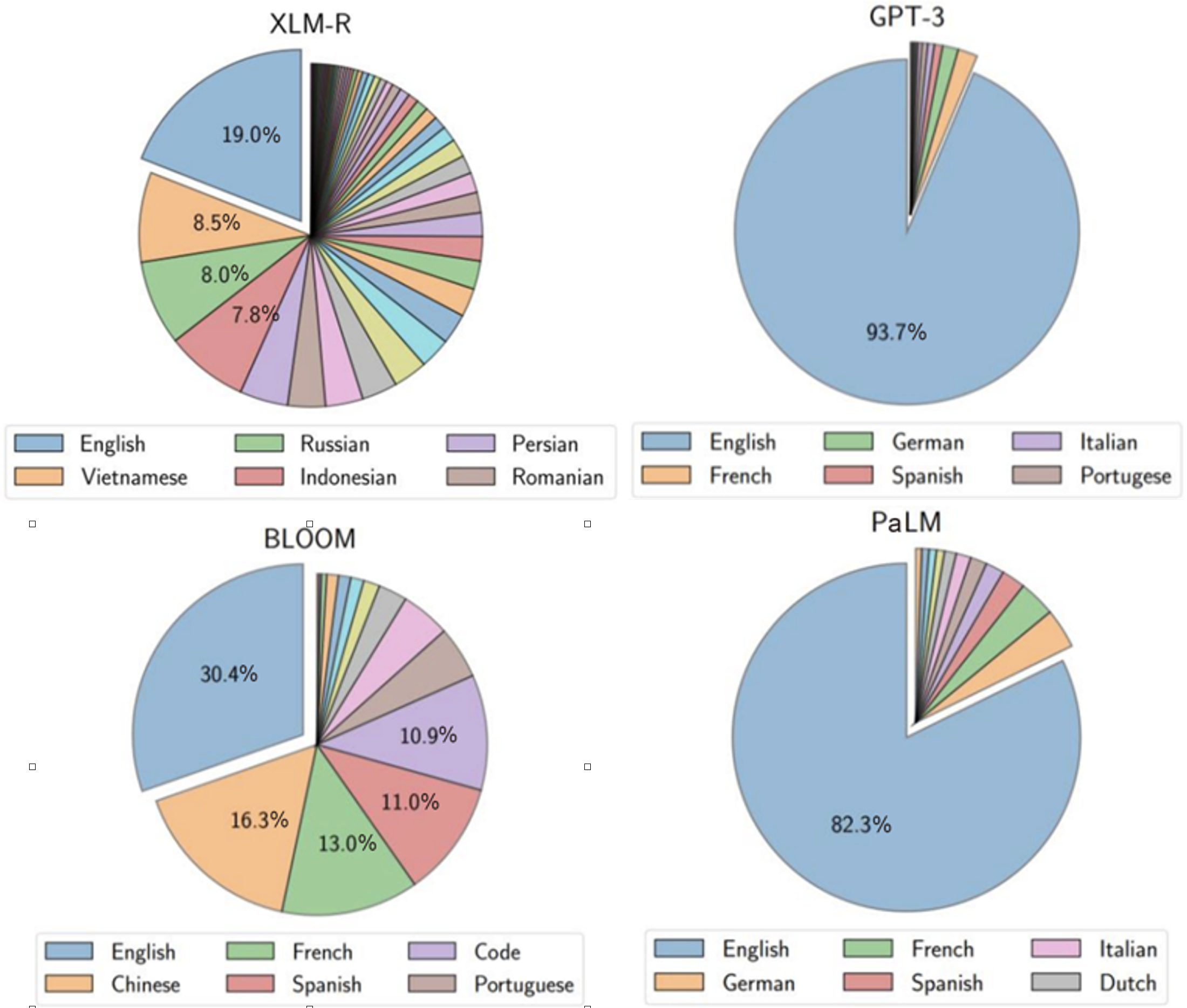}
    \caption{Proportion of main languages in common MLLMs}
    \label{fig:mllm-dist}
    \vspace{-.5cm}
\end{figure}

\textbf{Prefix LM Structure:}
The prefix LM is essentially still a decoder structure, but it changes the attention mechanism in the decoder structure. As shown in Figure~\ref{fig:lmsub3}, it uses bidirectional attention mechanisms for the tokens of the input sequence prefix (as shown in green in Figure~\ref{fig:lmsub3}), while other tokens use unidirectional attention mechanisms (as shown in orange in Figure~\ref{fig:lmsub3}). In this way, the prefix language model can use bidirectional attention mechanisms to collect as much language information as possible for the input sequence, while the output sequence uses an autoregressive method for one-by-one prediction, and shares parameters in both encoding and decoding processes. Compared with the previously mentioned encoder-decoder structure, this structure delegates the encoding work of the encoder to the decoder, saving 2 times the parameters and memory while ensuring similar computational complexity. Generally, we won't use this structure to train MLLMs from scratch, but will first train a decoder model, then transform the task into a corpus format that matches the input format of this model, and perform secondary training on the model to accelerate convergence speed. Common prefix language models include Llama2-chat, GLM-130B, and U-PaLM.

\begin{figure*}[t]
    \centering
    \includegraphics[width=\textwidth]{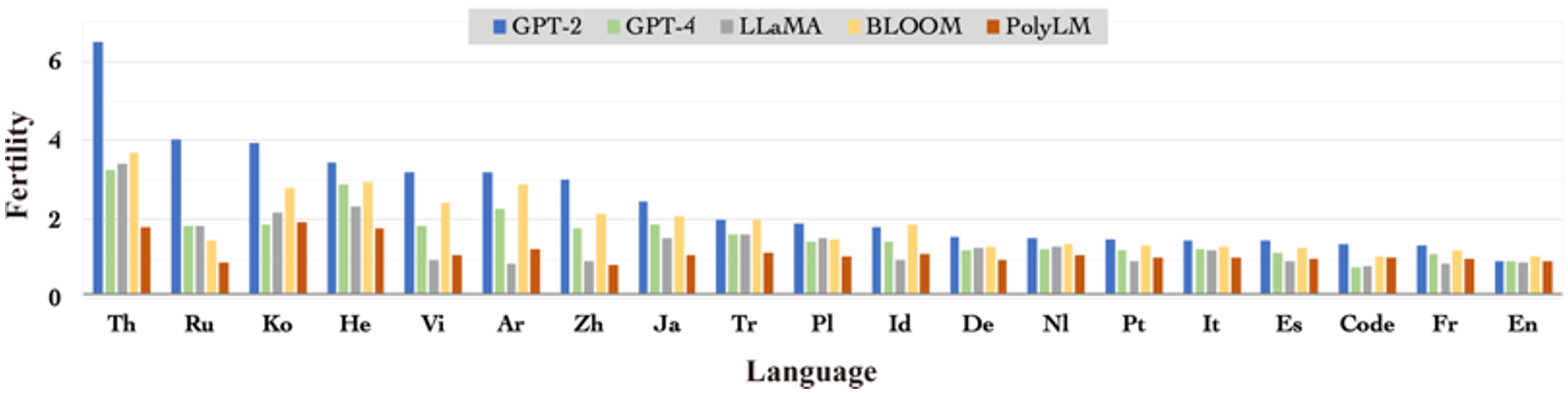}
    \caption{Comparison of perplexity between common MLLMs in PolyLM technical report and XLM-R}
    \label{fig:lang-fertility}
\end{figure*}

\subsection{Summary of Optimizing MLLMs}
This section will use Llama2 as an example to comprehensively explain how to use Llama2 to train a large model proficient in multilingual tasks from three aspects: pre-training corpus preparation, training data sampling, and training objectives and model fine-tuning, based on the technical directions introduced earlier.
\subsubsection{Pre-training Corpus Enhancement}
In Llama2's training corpus, as shown in Figure~\ref{fig:llama2-dist}, English corpus accounts for 89.7\%, other language corpora account for less than 2\% each, and there is about 8.38\% of programming language corpus. Although Llama2's technical report states that Llama2 has achieved performance comparable to GPT-3 on English datasets, it hasn't shown good performance on multilingual (such as Chinese) tasks~\cite{meta2023warning}. Even META has stated that Llama2 is not suitable for use in other languages. Therefore, if you want to train a MLLM based on Llama2, the first step is to collect more multilingual training corpora.

Figure~\ref{fig:mllm-dist} shows the proportion of main languages in common MLLMs. If we want to build a multilingual large model based on Llama2, the language proportion chosen should best refer to the ratios in Figure~\ref{fig:mllm-dist}. Generally, corpus enhancement can be divided into two categories: monolingual training corpus enhancement and multilingual training corpus enhancement.

\textbf{Monolingual training corpus enhancement:}
Generally, if we want to improve Llama2's performance on tasks in languages other than English, we need to collect corpora for specific languages and retrain Llama2 (such as Chinese Llama2). In addition to collecting corpora equivalent to English, we can also use the collected corpora to generate new tokens using encoding algorithms like BPE, supplement them to the original Llama2 dictionary, and then start model retraining. This can improve the model's performance on tasks in corresponding languages.

\textbf{Multilingual training corpus enhancement:}
The model enhanced with monolingual training corpus mentioned above, although it can improve capabilities in specific languages, still lacks multilingual capabilities and cannot solve the model's bias towards high-resource languages. A typical application of multilingual capability is machine translation. When we want to train a multilingual large model capable of various language translations, we still need to collect sufficient multilingual training corpora. Currently public multilingual datasets such as mC4, ROOT, and OPUS-100~\cite{zhang2020improving} can all be used as enhancement corpora for large models.

As mentioned in Section~\ref{sec:parallel}, large models trained with multilingual parallel sentence pairs generally perform better on multilingual downstream tasks compared to models trained only with monolingual or English-centric parallel sentence pairs. Therefore, when constructing multilingual training corpora, we recommend building multilingual parallel corpora and utilizing data augmentation to balance the number of mutual translation pairs, making them as similar as possible. For example, the amount of English-to-Chinese translation corpus should be similar to the Chinese-to-English translation corpus.

After completing the enhancement of training corpora, if we want to continue using new token enhancement or training our own tokenizer, it is recommended to use XLM-R's tokenizer as a benchmark and evaluate our tokenizer's fertility rate across different languages. XLM-R's dictionary deliberately includes support for low-resource language tokens, effectively addressing the high-resource language bias problem during pre-training model training. Therefore, if your tokenizer's fertility rate for different languages is similar to or lower than XLM-R's, it indicates that your model can effectively parse these language features and improve efficiency during inference and training.

Figure~\ref{fig:lang-fertility} shows the fertility rate comparison between common MLLMs and XLM-R from the PolyLM technical report~\cite{wei2023polylm}. As can be observed, PolyLM's tokenizer provides excellent support for multiple languages. Interested readers can also use their training corpora and partial sampling to retrain or fine-tune Llama2 for their specific tasks.

\begin{figure*}[t]
    \centering
    \includegraphics[width=\textwidth]{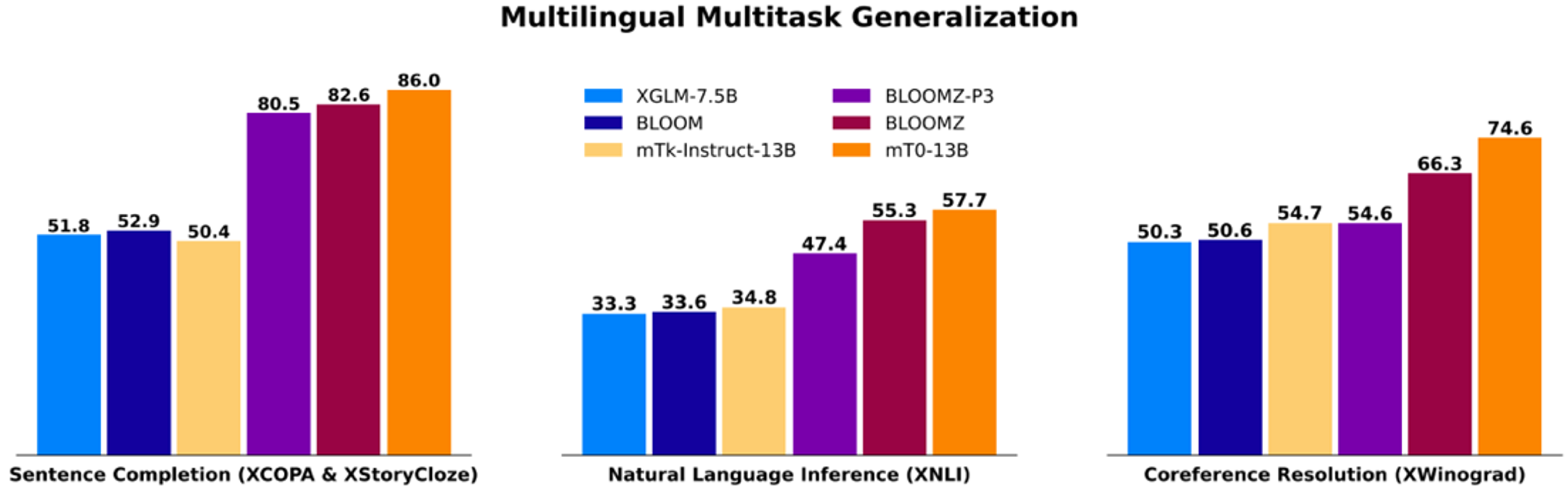}
    \caption{Zero-shot multilingual task performance chart (all prompts in English)}
    \label{fig:zeroshot-mlt}
\end{figure*}

\subsubsection{Tokenizer Selection}
Generally, we will use the BPE algorithm in SentencePiece for training. It should be noted that most large models (including Llama2) will split characters representing numbers into separate digits during training to enhance the ability to answer mathematical problems. For different languages, there may be different expressions from Arabic numerals. Although the amount of data may not be large, careful readers can also convert them to Arabic numerals for unified processing. At the same time, Llama2 will also decompose unrecognized characters into bytes using the BBPE mentioned above to improve coverage of rare characters (such as emojis), which readers can directly reuse.

\subsubsection{Training Language Sampling and Training Objectives}
After we prepare the corpus and are ready to retrain Llama2, we still need to pay attention to the training language sampling strategy mentioned earlier, including temperature-based sampling strategy and Unimax method. If you choose the temperature-based sampling strategy, it is recommended to use a temperature coefficient of 3.33, which is also used by XLM-R.

In addition to language sampling, you also need to choose appropriate model training structures and training objectives according to your training purpose. If you want to train a multilingual large model based on Llama2, you can use the decoder and NTP technology for training; if you only want to fine-tune for multilingual tasks using existing corpora, you can choose to use the prefix language model and match it with different decoding objectives. For example, for multilingual instruction fine-tuning, you can choose Specific-Denoising as the training objective, only calculating the loss for the answer; if you want to train a Chatbot supporting multi-turn interaction, you need to use Extreme-Denoising as the training objective, because we only need to calculate the loss of each round of Chatbot's response, without considering the loss of user questions, although our model will see and pay attention to all context information in the Transformer layer.

\subsubsection{Fine-tuning MLLMs}
When we have clarified our goals (retraining or fine-tuning goals), prepared the corpus, tokenizer dictionary, training objectives, and hyperparameters, we can proceed with model retraining or fine-tuning for Llama2.

\textbf{Multilingual retraining:}
Generally, model retraining will use common decoder structures, and the training objective is also to predict the next character. However, in multilingual training, to avoid the model overfitting high-resource languages and thus hindering the learning of low-resource languages, curriculum learning~\cite{bengio2009curriculum} methods are usually adopted for multilingual large model pre-training. This approach aims to transfer some general knowledge from high-resource languages to low-resource languages while retaining the model's excellent capabilities in high-resource languages. The general approach is to divide the training into two stages: the first stage focuses on high-resource training corpora, trying to let the model learn more general language knowledge; the second stage expands the proportion of low-resource training corpora, hoping the model can enhance multilingual capabilities. For instance, past work proposed curriculum learning methods for BigTranslate~\cite{yang2023bigtrans} and PolyLM~\cite{wei2023polylm}, respectively.

\textbf{Multilingual instruction fine-tuning:}
Usually, if we don't have enough GPU and corpus resources for multilingual large model pre-training, but still hope that Llama2 can perform well on tasks in different languages, we can choose multilingual instruction fine-tuning. Its model structure and training objectives have been introduced earlier. In the research paper of BLOOMZ~\cite{muennighoff2022crosslingual}, it was found that using only English tasks to fine-tune MLLMs, the resulting model BLOOMZ-P3 can obtain better results than the original model (BLOOM); if using multilingual tasks and English instructions to fine-tune BLOOM, then BLOOMZ can get better results than before. Given that the Llama2 we are using is primarily English-based, it is recommended to use the latter for model fine-tuning to improve Llama2's performance on multilingual tasks at minimal cost.

Zero-shot multilingual multitask performance is shown in Figure~\ref{fig:zeroshot-mlt}. When constructing the dataset for instruction fine-tuning, it is recommended to use higher quality data for fine-tuning. According to the LIMA paper~\cite{zhou2023lima}, the quality of fine-tuning data is more important than quantity.

\begin{figure}[t]
    \centering
    \includegraphics[width=.48\textwidth]{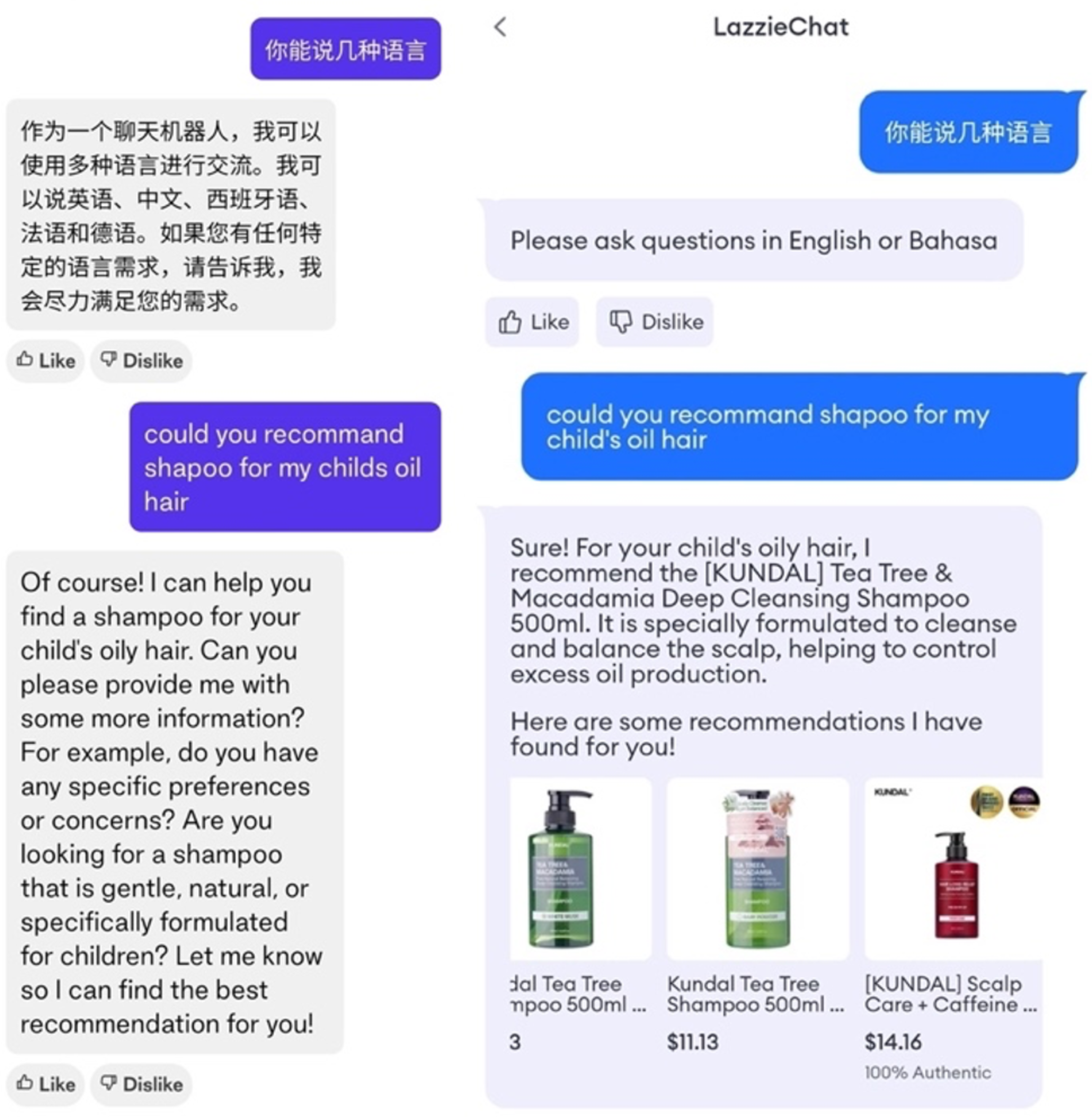}
    \caption{Illustration of Shopify bot and LazzieChat bot}
    \label{fig:shopify-lazziechat}
\end{figure}

\begin{figure*}[t]
    \centering
    \includegraphics[width=.8\textwidth]{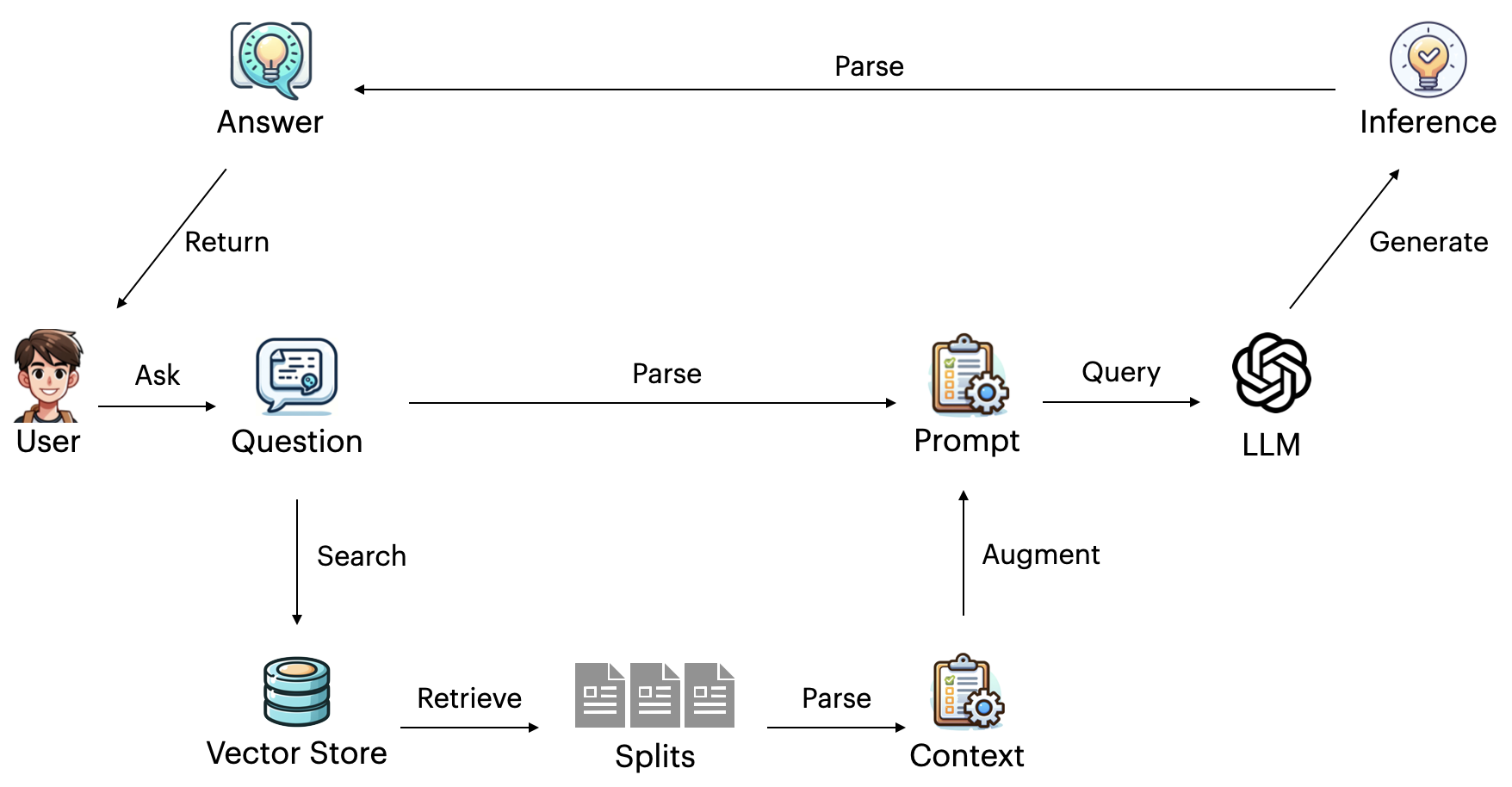}
    \caption{Retrieval-Augmented Generation process diagram}
    \label{fig:rag}
\end{figure*}

\section{Industrial Applications of MLLMs}
The emergence of large language models, i.e., generative AI, allows people to interact with machines using natural language as instructions, while obtaining desired answers through various APIs and operations. Its appearance is reshaping many industries and industrial applications. This section will focus on introducing several typical industrial applications that require MLLMs for interaction. These industrial applications leverage MLLMs to enhance user experience in related fields or improve operational efficiency.

\subsection{Intelligent Customer Service}
One of the most typical industrial applications of MLLMs is intelligent customer service. The inherent multi-turn dialogue capability of large models allows them to more naturally answer pre-sales and after-sales questions raised by users, improving enterprise service efficiency. At the same time, large models can also be used to recognize users' emotions, provide comfort or escalate issues for different users, improving user satisfaction and reducing various public relations risks.

For pre-sales questions, intelligent customer service is more like a shopping guide robot, which will ask various counter-questions based on user needs to collect information, and finally recommend products that users might be interested in to increase user conversion rates. Recently, Southeast Asian e-commerce platform Lazada and Shopify have respectively launched their own LazzieChat bot and Shopify bot based on OpenAI's ChatGPT, as shown in Figure~\ref{fig:shopify-lazziechat}. They can recommend products based on user needs, with LazzieChat bot mainly supporting English and Indonesian, while Shopify bot supports Chinese and more European languages.

For answering after-sales FAQs, because it involves professional domain knowledge, such as inconsistent policies on different platforms, and the timeliness of knowledge is relatively high, the Retrieval-Augmented Generation (RAG) method~\cite{liu2024lara,lewis2020retrieval} is generally adopted. This can utilize external new knowledge and to some extent solve the hallucination problem of large models. The specific process is shown in Figure~\ref{fig:rag}. Generally, a vector database will be constructed first for domain knowledge, then relevant documents will be retrieved from the vector database using the user's question, and then the question, documents, and corresponding prompts will be given to the large model to complete the answer generation and reply to the user. The whole process is very similar to the previous reading comprehension and document QA tasks.

In addition to simple document QA, sometimes intelligent customer service also needs to conduct multi-turn task-oriented dialogues~\cite{liu2024mint}, which is Taskbot. In these tasks, large models not only need to have multi-turn dialogue capabilities but also need to have tool use (such as API calling) capabilities to query slot information required for tasks, and the answers will also change dynamically based on the queried slot information. For example, if a user asks when their order will arrive, if their order status is already returned, we need to explain to the user that their order is in a returned status and cannot be received anymore, rather than giving a generic answer. Currently, research in this area mainly focuses on using large models to convert user instructions into actions or code that machines can understand, and then execute them in specific environments, i.e., LLM-AGENT~\cite{weng2023agent}. Interested readers can look at projects like AutoGPT or XLANG~\cite{xlang2023}, which develop LLM-AGENTs that act as bridges between natural language and specific instructions (API calls or action sequences, such as configured dialogue processes). Through rounds of interaction with the environment and humans, large models can collect sufficient context information to complete tasks accurately and effectively, extending and expanding user intentions~\cite{wechat2023translation}.

\subsection{Search Engines}
\begin{figure}[t]
    \centering
    \includegraphics[width=.48\textwidth]{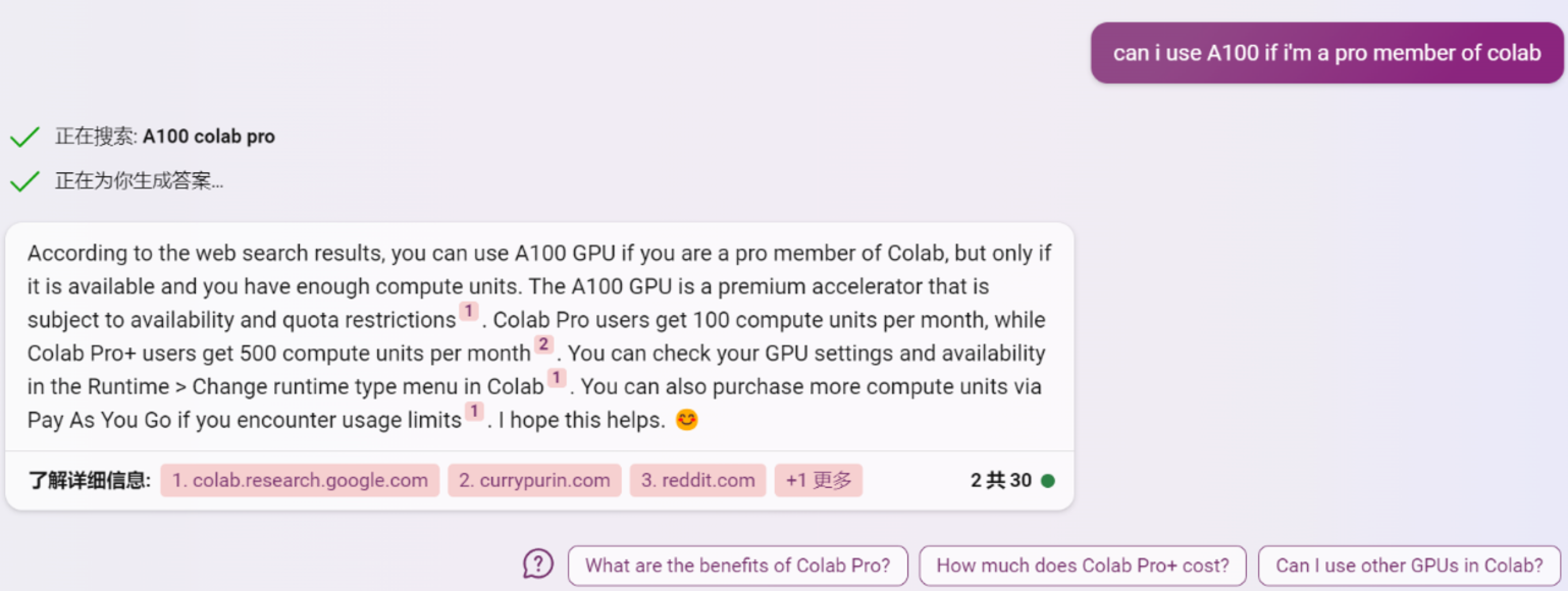}
    \caption{Demonstration of Bing Chat system}
    \label{fig:bing}
\end{figure}

In addition to intelligent customer service, companies like Google and Microsoft's Bing are also using RAG technology and MLLMs to provide web search services. Search engines that use large models can provide more concise responses, although at the cost of potentially inaccurate answers. Figure~\ref{fig:bing} shows a demonstration of the Bing Chat system. To prevent inaccurate information from misleading users, it also marks the sources of different responses to help users better find answers.

\subsection{Machine Translation}
\begin{figure}[t]
    \centering
    \includegraphics[width=.48\textwidth]{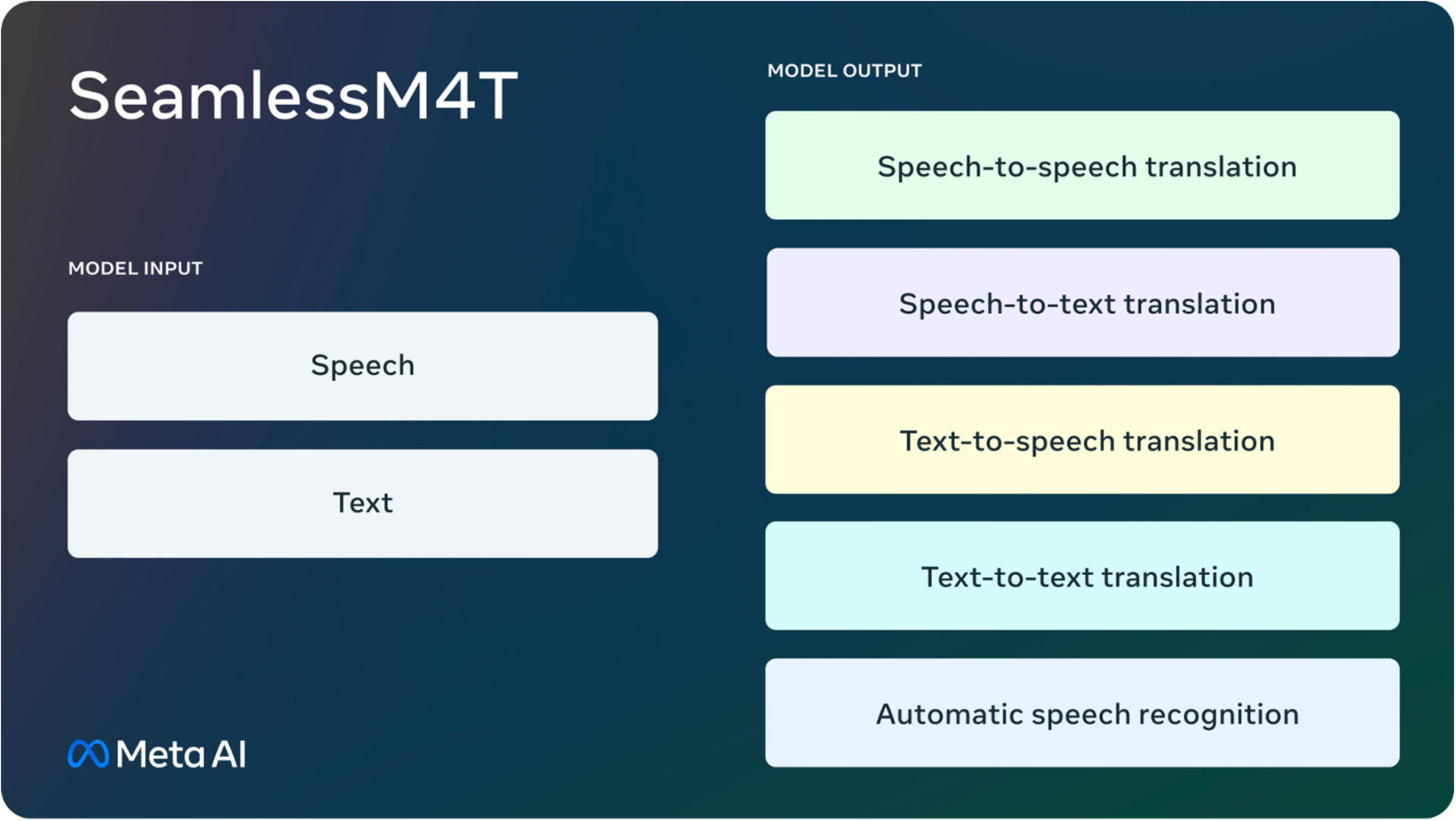}
    \caption{Overview of Meta's Seamless M4T}
    \label{fig:seamlessm4t}
\end{figure}

Machine translation is a scenario that naturally requires the use of MLLMs. The current translation effect of MLLMs is basically consistent with that of original machine translation models for high-resource languages, although MLLMs still have a significant gap with Google's translation for low-resource language pairs. However, they can basically meet the translation needs for common language pairs. At the same time, because MLLMs can better understand context, they may have better effects on the translation of some conference and business documents. Currently, more mature applications include Meta AI's Seamless M4T~\cite{barrault2023seamlessm4t}, as shown in Figure~\ref{fig:seamlessm4t}, and BigTranslate, a multilingual large model trained by the Institute of Automation, Chinese Academy of Sciences using Llama as the base model.

\section{Conclusion}

This survey provides a comprehensive overview of multilingual large language models (MLLMs), highlighting their crucial role in advancing natural language processing beyond English-centric approaches. The development of MLLMs has emerged as a critical necessity for bridging the digital language divide, particularly given that over 7,000 languages are spoken worldwide, yet only a small fraction are well-represented in current AI technologies. This disparity is especially pronounced for the 88.38\% of languages categorized as "left-behind" or low-resource, despite these languages serving over a billion speakers globally.

The field faces several significant technical challenges in developing effective MLLMs. The "curse of multilinguality" presents a fundamental challenge in balancing model capacity across multiple languages, while tokenization complexities for morphologically rich languages create additional hurdles. The uneven distribution of training data across languages and higher inference costs for non-Latin script languages further complicate the development landscape.

To address these challenges, the field has developed several innovative methodological approaches. Sophisticated sampling strategies, including temperature-based sampling and Unimax, have emerged to better balance language representation in training. Various model architectures, including decoder-only, encoder-decoder, and prefix language models, have been explored to optimize multilingual performance. Advanced training objectives combining Regular-Denoising, Specific-Denoising, and Extreme-Denoising have shown promise in improving model capabilities, while curriculum learning approaches have helped balance the treatment of high and low-resource languages.

The evaluation and benchmarking of MLLMs has matured significantly with the emergence of comprehensive frameworks like XTREME, MEGA, and BUFFET. These frameworks have enabled more robust assessment of multilingual capabilities across diverse tasks and languages, providing crucial insights for continued development. In parallel, MLLMs have found practical applications in various industrial contexts, from intelligent customer service systems and multilingual search engines to advanced machine translation systems and task-oriented dialogue systems.

Looking ahead, several research directions show particular promise for advancing the field. The development of more efficient tokenization strategies for low-resource languages remains a crucial area for improvement, as does the refinement of cross-lingual knowledge transfer methods. Enhanced techniques for reducing the computational costs of multilingual models will be essential for broader adoption, while better integration of linguistic and cultural knowledge in model training could improve performance and cultural sensitivity.

The field of MLLMs represents a crucial step toward more inclusive and equitable AI technologies. While significant progress has been made, continued research and development are essential to address remaining challenges and further expand the linguistic capabilities of these models. Future work should focus on improving support for low-resource languages, reducing computational requirements, and developing more sophisticated evaluation metrics that consider both technical performance and cultural sensitivity. These advances will be crucial in realizing the full potential of MLLMs and ensuring that AI technology benefits speakers of all languages, ultimately working toward a more linguistically inclusive digital future.

\bibliographystyle{IEEEtran}
\bibliography{ref}

\begin{thebibliography}{10}
\providecommand{\url}[1]{#1}
\csname url@samestyle\endcsname
\providecommand{\newblock}{\relax}
\providecommand{\bibinfo}[2]{#2}
\providecommand{\BIBentrySTDinterwordspacing}{\spaceskip=0pt\relax}
\providecommand{\BIBentryALTinterwordstretchfactor}{4}
\providecommand{\BIBentryALTinterwordspacing}{\spaceskip=\fontdimen2\font plus
\BIBentryALTinterwordstretchfactor\fontdimen3\font minus \fontdimen4\font\relax}
\providecommand{\BIBforeignlanguage}[2]{{%
\expandafter\ifx\csname l@#1\endcsname\relax
\typeout{** WARNING: IEEEtran.bst: No hyphenation pattern has been}%
\typeout{** loaded for the language `#1'. Using the pattern for}%
\typeout{** the default language instead.}%
\else
\language=\csname l@#1\endcsname
\fi
#2}}
\providecommand{\BIBdecl}{\relax}
\BIBdecl

\bibitem{ethnologue}
Ethnologue, ``Ethnologue: Languages of the world,'' \url{https://www.ethnologue.com/}, 2024.

\bibitem{ruder_nlp}
S.~Ruder, ``Natural language processing beyond english,'' \url{https://www.ruder.io/nlp-beyond-english/?ref=ruder.io}, 2024.

\bibitem{xu2024survey}
Y.~Xu, L.~Hu, J.~Zhao, Z.~Qiu, Y.~Ye, and H.~Gu, ``A survey on multilingual large language models: Corpora, alignment, and bias,'' \emph{arXiv preprint arXiv:2404.00929}, 2024.

\bibitem{qin2024multilingual}
L.~Qin, Q.~Chen, Y.~Zhou, Z.~Chen, Y.~Li, L.~Liao, M.~Li, W.~Che, and P.~S. Yu, ``Multilingual large language model: A survey of resources, taxonomy and frontiers,'' \emph{arXiv preprint arXiv:2404.04925}, 2024.

\bibitem{huang2024survey}
K.~Huang, F.~Mo, H.~Li, Y.~Li, Y.~Zhang, W.~Yi, Y.~Mao, J.~Liu, Y.~Xu, J.~Xu \emph{et~al.}, ``A survey on large language models with multilingualism: Recent advances and new frontiers,'' \emph{arXiv preprint arXiv:2405.10936}, 2024.

\bibitem{qz2017google}
Q.~Media, ``Google translate's gender bias: Pairs "he" with hardworking and "she" with lazy,'' 2017.

\bibitem{alexa_accent}
T.~W. Post, ``Alexa, siri, and other voice assistants frequently fail to understand people with non-american accents,'' \url{https://www.washingtonpost.com/graphics/2018/business/alexa-does-not-understand-your-accent/?ref=ruder.io}, accessed: 2024-10-18.

\bibitem{antoun2020arabert}
W.~Antoun, F.~Baly, and H.~Hajj, ``Arabert: Transformer-based model for arabic language understanding,'' \emph{arXiv preprint arXiv:2003.00104}, 2020.

\bibitem{devlin2018bert}
J.~Devlin, ``Bert: Pre-training of deep bidirectional transformers for language understanding,'' \emph{arXiv preprint arXiv:1810.04805}, 2018.

\bibitem{virtanen2019multilingual}
A.~Virtanen, J.~Kanerva, R.~Ilo, J.~Luoma, J.~Luotolahti, T.~Salakoski, F.~Ginter, and S.~Pyysalo, ``Multilingual is not enough: Bert for finnish,'' \emph{arXiv preprint arXiv:1912.07076}, 2019.

\bibitem{wilie2020indonlu}
B.~Wilie, K.~Vincentio, G.~I. Winata, S.~Cahyawijaya, X.~Li, Z.~Y. Lim, S.~Soleman, R.~Mahendra, P.~Fung, S.~Bahar \emph{et~al.}, ``Indonlu: Benchmark and resources for evaluating indonesian natural language understanding,'' \emph{arXiv preprint arXiv:2009.05387}, 2020.

\bibitem{cl_to_bert_japanese}
\BIBentryALTinterwordspacing
{Tohoku University NLP Lab}, ``Japanese bert pretrained models,'' \url{https://github.com/cl-tohoku/bert-japanese}, 2021, gitHub repository. [Online]. Available: \url{https://github.com/cl-tohoku/bert-japanese}
\BIBentrySTDinterwordspacing

\bibitem{lee2020study}
C.~H. Lee, Y.~J. Lee, and D.~H. Lee, ``A study of fine tuning pre-trained korean bert for question answering performance development,'' \emph{Journal of Information Technology Services}, vol.~19, no.~5, pp. 83--91, 2020.

\bibitem{kuratov2019adaptation}
Y.~Kuratov and M.~Arkhipov, ``Adaptation of deep bidirectional multilingual transformers for russian language,'' \emph{arXiv preprint arXiv:1905.07213}, 2019.

\bibitem{schweter2020berturk}
S.~Schweter, ``Berturk-bert models for turkish,'' \emph{Zenodo}, vol. 2020, p. 3770924, 2020.

\bibitem{zhihu2022ai}
Zhihu, ``Ai article on zhihu,'' \url{https://zhuanlan.zhihu.com/p/525515142}, 2022, accessed: 2024-10-22.

\bibitem{chelliah2001sinotibetan}
S.~L. Chelliah, ``Sino-tibetan linguistics,'' \url{https://folklife-media.si.edu/docs/folklife/stlrmw/sino-tibetan-linguistics-chelliah-01.pdf}, 2001, accessed: 2024-10-22.

\bibitem{wals2024}
M.~P.~I. for Evolutionary~Anthropology, ``The world atlas of language structures (wals),'' \url{https://wals.info/}, 2024, accessed: 2024-10-22.

\bibitem{joshi2020state}
P.~Joshi \emph{et~al.}, ``The state and fate of linguistic diversity and inclusion in the nlp world,'' \emph{arXiv preprint arXiv:2004.09095}, 2020.

\bibitem{artetxe2020call}
M.~Artetxe \emph{et~al.}, ``A call for more rigor in unsupervised cross-lingual learning,'' \emph{arXiv preprint arXiv:2004.14958}, 2020.

\bibitem{liu2022title2vec}
J.~Liu, Y.~C. Ng, Z.~Gui, T.~Singhal, L.~T.~M. Blessing, K.~L. Wood, and K.~H. Lim, ``Title2vec: a contextual job title embedding for occupational named entity recognition and other applications,'' \emph{Journal of Big Data}, vol.~9, no.~1, pp. 1--16, 2022.

\bibitem{liu2024spatial}
J.~Liu, J.~Albrethsen, L.~Goh, D.~Yau, and K.~H. Lim, ``Spatial-temporal graph representation learning for tactical networks future state prediction,'' in \emph{IJCNN}, 2024.

\bibitem{bender2011language_independence}
E.~M. Bender, ``On achieving and evaluating language-independence in nlp,'' \emph{Linguistic Issues in Language Technology}, vol.~6, 2011.

\bibitem{tsarfaty2020spmrl}
R.~Tsarfaty \emph{et~al.}, ``From spmrl to nmrl: What did we learn (and unlearn) in a decade of parsing morphologically-rich languages (mrls)?'' \emph{arXiv preprint arXiv:2005.01330}, 2020.

\bibitem{vania2017morphology}
C.~Vania and A.~Lopez, ``From characters to words to in between: Do we capture morphology?'' \emph{arXiv preprint arXiv:1704.08352}, 2017.

\bibitem{zeng2023soft}
J.~Zeng \emph{et~al.}, ``Soft language clustering for multilingual model pre-training,'' \emph{arXiv preprint arXiv:2306.07610}, 2023.

\bibitem{muennighoff2022crosslingual}
N.~Muennighoff \emph{et~al.}, ``Crosslingual generalization through multitask finetuning,'' \emph{arXiv preprint arXiv:2211.01786}, 2022.

\bibitem{scao2022bloom}
T.~L. Scao \emph{et~al.}, ``Bloom: A 176b-parameter open-access multilingual language model,'' \emph{arXiv preprint arXiv:2211.05100}, 2022.

\bibitem{xue2020mt5}
L.~Xue \emph{et~al.}, ``mt5: A massively multilingual pre-trained text-to-text transformer,'' \emph{arXiv preprint arXiv:2010.11934}, 2020.

\bibitem{conneau2019unsupervised}
A.~Conneau \emph{et~al.}, ``Unsupervised cross-lingual representation learning at scale,'' \emph{arXiv preprint arXiv:1911.02116}, 2019.

\bibitem{goyal2021larger}
N.~Goyal \emph{et~al.}, ``Larger-scale transformers for multilingual masked language modeling,'' \emph{arXiv preprint arXiv:2105.00572}, 2021.

\bibitem{peft}
\BIBentryALTinterwordspacing
H.~Face, ``Parameter-efficient fine-tuning,'' 2023, accessed: 2024-10-17. [Online]. Available: \url{https://github.com/huggingface/peft}
\BIBentrySTDinterwordspacing

\bibitem{nopss}
\BIBentryALTinterwordspacing
N.~P. Service and S.~S.~B. of~China, ``Notice on social security reform,'' 2019, accessed: 2024-10-17. [Online]. Available: \url{http://www.nopss.gov.cn/n1/2019/1121/c219470-31468057.html}
\BIBentrySTDinterwordspacing

\bibitem{koto2020indolem}
F.~Koto \emph{et~al.}, ``Indolem and indobert: A benchmark dataset and pre-trained language model for indonesian nlp,'' \emph{arXiv preprint arXiv:2011.00677}, 2020.

\bibitem{nguyen2020phobert}
D.~Q. Nguyen and A.~T. Nguyen, ``Phobert: Pre-trained language models for vietnamese,'' \emph{arXiv preprint arXiv:2003.00744}, 2020.

\bibitem{lowphansirikul2021wangchanberta}
L.~Lowphansirikul \emph{et~al.}, ``Wangchanberta: Pretraining transformer-based thai language models,'' \emph{arXiv preprint arXiv:2101.09635}, 2021.

\bibitem{huang2023not}
H.~Huang \emph{et~al.}, ``Not all languages are created equal in llms: Improving multilingual capability by cross-lingual-thought prompting,'' \emph{arXiv preprint arXiv:2305.07004}, 2023.

\bibitem{zhao2023survey}
W.~X. Zhao \emph{et~al.}, ``A survey of large language models,'' \emph{arXiv preprint arXiv:2303.18223}, 2023.

\bibitem{zhu2015aligning}
Y.~Zhu, R.~Kiros, R.~S. Zemel, R.~Salakhutdinov, R.~Urtasun, A.~Torralba, and S.~Fidler, ``Aligning books and movies: Towards story-like visual explanations by watching movies and reading books,'' in \emph{2015 IEEE International Conference on Computer Vision (ICCV)}.\hskip 1em plus 0.5em minus 0.4em\relax IEEE, 2015, pp. 19--27.

\bibitem{project_gutenberg}
Gutenberg, ``Project gutenberg,'' \url{https://www.gutenberg.org/}, accessed: 2024-10-18.

\bibitem{raffel2020exploring}
C.~Raffel \emph{et~al.}, ``Exploring the limits of transfer learning with a unified text-to-text transformer,'' \emph{The Journal of Machine Learning Research}, vol.~21, no.~1, pp. 5485--5551, 2020.

\bibitem{abadji2022towards}
J.~Abadji \emph{et~al.}, ``Towards a cleaner document-oriented multilingual crawled corpus,'' \emph{arXiv preprint arXiv:2201.06642}, 2022.

\bibitem{el-kishky2019ccaligned}
A.~El-Kishky \emph{et~al.}, ``Ccaligned: A massive collection of cross-lingual web-document pairs,'' \emph{arXiv preprint arXiv:1911.06154}, 2019.

\bibitem{baumgartner2020pushshift}
J.~Baumgartner \emph{et~al.}, ``The pushshift reddit dataset,'' in \emph{Proceedings of the International AAAI Conference on Web and Social Media}, vol.~14, 2020.

\bibitem{schwenk2019wikimatrix}
H.~Schwenk \emph{et~al.}, ``Wikimatrix: Mining 135m parallel sentences in 1620 language pairs from wikipedia,'' \emph{arXiv preprint arXiv:1907.05791}, 2019.

\bibitem{google_bigquery}
Google, ``Google cloud bigquery public datasets,'' \url{https://cloud.google.com/bigquery/public-data}, accessed: 2024-10-18.

\bibitem{nijkamp2022codegen}
E.~Nijkamp \emph{et~al.}, ``Codegen: An open large language model for code with multi-turn program synthesis,'' \emph{arXiv preprint arXiv:2203.13474}, 2022.

\bibitem{liu2021crisisbert}
J.~Liu, T.~Singhal, L.~T. Blessing, K.~L. Wood, and K.~H. LIM, ``Crisisbert: a robust transformer for crisis classification and contextual crisis embedding,'' in \emph{Proceedings of the 32nd ACM Conference on Hypertext and Social Media}, 2021, pp. 133--141.

\bibitem{li2023transformer}
M.~Li, K.~H. Lim, T.~Guo, and J.~Liu, ``A transformer-based framework for poi-level social post geolocation,'' in \emph{ECIR}.\hskip 1em plus 0.5em minus 0.4em\relax Springer Nature Switzerland, 2023, pp. 588--604.

\bibitem{ahuja2023mega}
K.~Ahuja \emph{et~al.}, ``Mega: Multilingual evaluation of generative ai,'' \emph{arXiv preprint arXiv:2303.12528}, 2023.

\bibitem{asai2023buffet}
A.~Asai \emph{et~al.}, ``Buffet: Benchmarking large language models for few-shot cross-lingual transfer,'' \emph{arXiv preprint arXiv:2305.14857}, 2023.

\bibitem{ding2021globalwoz}
B.~Ding \emph{et~al.}, ``Globalwoz: Globalizing multiwoz to develop multilingual task-oriented dialogue systems,'' \emph{arXiv preprint arXiv:2110.07679}, 2021.

\bibitem{moradshahi2023xrisa}
M.~Moradshahi \emph{et~al.}, ``X-risawoz: High-quality end-to-end multilingual dialogue datasets and few-shot agents,'' \emph{arXiv preprint arXiv:2306.17674}, 2023.

\bibitem{huggingface2023nlp}
Huggingface, ``Nlp course: Chapter 6.5,'' \url{https://huggingface.co/learn/nlp-course/chapter6/5?fw=pt}, 2023.

\bibitem{sentencepiece2023}
Google, ``Sentencepiece,'' \url{https://github.com/google/sentencepiece}, 2023.

\bibitem{wang2020neural}
C.~Wang, K.~Cho, and J.~Gu, ``Neural machine translation with byte-level subwords,'' in \emph{Proceedings of the AAAI conference on artificial intelligence}, vol. 34-05, 2020, pp. 9154--9160.

\bibitem{ahia2023tokenization}
O.~Ahia \emph{et~al.}, ``Do all languages cost the same? tokenization in the era of commercial language models,'' \emph{arXiv preprint arXiv:2305.13707}, 2023.

\bibitem{fujii2023tokenizers}
T.~Fujii \emph{et~al.}, ``How do different tokenizers perform on downstream tasks in scriptio continua languages?: A case study in japanese,'' \emph{arXiv preprint arXiv:2306.09572}, 2023.

\bibitem{chung2020improving}
H.~W. Chung \emph{et~al.}, ``Improving multilingual models with language-clustered vocabularies,'' \emph{arXiv preprint arXiv:2010.12777}, 2020.

\bibitem{chung2023unimax}
H.~W. Chung, N.~Constant, X.~Garcia, A.~Roberts, Y.~Tay, S.~Narang, and O.~Firat, ``Unimax: Fairer and more effective language sampling for large-scale multilingual pretraining,'' \emph{arXiv preprint arXiv:2304.09151}, 2023.

\bibitem{paul2013pivot}
M.~Paul, A.~Finch, and E.~Sumita, ``How to choose the best pivot language for automatic translation of low-resource languages,'' \emph{ACM Transactions on Asian Language Information Processing (TALIP)}, vol.~12, no.~4, pp. 1--17, 2013.

\bibitem{bing2022turing}
Bing, ``Microsoft turing universal language representation model, t-ulrv6, tops both xtreme and glue leaderboard,'' 2022.

\bibitem{tay2022ul2}
Y.~Tay, V.~Q. Tran, S.~Ruder, J.~Gupta, H.~W. Chung, D.~Bahri, Z.~Qin, S.~Baumgartner, C.~Yu, and J.~Dean, ``Ul2: Unifying language learning paradigms,'' \emph{arXiv preprint arXiv:2205.05131}, 2022.

\bibitem{touvron2023llama2}
H.~Touvron, T.~Lavril, G.~Izacard, X.~Martinet, M.-A. Lachaux, T.~Lacroix, B.~Rozi{\`e}re, N.~Goyal, E.~Hambro, F.~Azhar \emph{et~al.}, ``Llama 2: Open foundation and fine-tuned chat models,'' \emph{arXiv preprint arXiv:2307.09288}, 2023.

\bibitem{anil2023palm2}
R.~Anil, A.~M. Dai, O.~Firat, M.~Johnson, D.~Lepikhin, M.~Nicolas, A.~Passos, S.~Shakeri, E.~Taropa, Y.~Tanaka \emph{et~al.}, ``Palm 2 technical report,'' \emph{arXiv preprint arXiv:2305.10403}, 2023.

\bibitem{vaswani2017attention}
A.~Vaswani \emph{et~al.}, ``Attention is all you need,'' in \emph{Advances in neural information processing systems}, vol.~30, 2017.

\bibitem{tay2022transcending}
Y.~Tay \emph{et~al.}, ``Transcending scaling laws with 0.1\% extra compute,'' \emph{arXiv preprint arXiv:2210.11399}, 2022.

\bibitem{chung2022scaling}
H.~W. Chung \emph{et~al.}, ``Scaling instruction-finetuned language models,'' \emph{arXiv preprint arXiv:2210.11416}, 2022.

\bibitem{meta2023warning}
\BIBentryALTinterwordspacing
{Slator}, ``Meta warns large language model may not be suitable for non-english use,'' 2023. [Online]. Available: \url{https://slator.com/meta-warns-large-language-model-may-not-be-suitable-non-english-use/}
\BIBentrySTDinterwordspacing

\bibitem{zhang2020improving}
B.~Zhang \emph{et~al.}, ``Improving massively multilingual neural machine translation and zero-shot translation,'' \emph{arXiv preprint arXiv:2004.11867}, 2020.

\bibitem{wei2023polylm}
X.~Wei \emph{et~al.}, ``Polylm: An open source polyglot large language model,'' \emph{arXiv preprint arXiv:2307.06018}, 2023.

\bibitem{bengio2009curriculum}
Y.~Bengio \emph{et~al.}, ``Curriculum learning,'' in \emph{Proceedings of the 26th annual international conference on machine learning}, 2009.

\bibitem{yang2023bigtrans}
W.~Yang \emph{et~al.}, ``Bigtrans: Augmenting large language models with multilingual translation capability over 100 languages,'' \emph{arXiv preprint arXiv:2305.18098}, 2023.

\bibitem{zhou2023lima}
C.~Zhou \emph{et~al.}, ``Lima: Less is more for alignment,'' \emph{arXiv preprint arXiv:2305.11206}, 2023.

\bibitem{liu2024lara}
J.~Liu, Y.~K. Tan, B.~Fu, and K.~H. Lim, ``Lara: Linguistic-adaptive retrieval-augmentation for multi-turn intent classification,'' in \emph{EMNLP}, 2024.

\bibitem{lewis2020retrieval}
P.~Lewis \emph{et~al.}, ``Retrieval-augmented generation for knowledge-intensive nlp tasks,'' in \emph{Advances in Neural Information Processing Systems}, vol.~33, 2020, pp. 9459--9474.

\bibitem{liu2024mint}
J.~Liu, Y.~K. Tan, B.~Fu, and K.~H. Lim, ``Intent-aware dialogue generation and multi-task contrastive learning for multi-turn intent classification,'' in \emph{ArXiv}, 2024.

\bibitem{weng2023agent}
\BIBentryALTinterwordspacing
L.~Weng, ``Thoughts on ai agents,'' 2023. [Online]. Available: \url{https://lilianweng.github.io/posts/2023-06-23-agent/}
\BIBentrySTDinterwordspacing

\bibitem{xlang2023}
\BIBentryALTinterwordspacing
{xLang}, ``{xLang},'' 2023. [Online]. Available: \url{https://github.com/xlang-ai/xlang}
\BIBentrySTDinterwordspacing

\bibitem{wechat2023translation}
\BIBentryALTinterwordspacing
{Xi, Xiaoyao}, ``Is chatgpt a good translator?'' 2023. [Online]. Available: \url{https://mp.weixin.qq.com/s/diK88UmXWGKqP-6ZhXQbBg}
\BIBentrySTDinterwordspacing

\bibitem{barrault2023seamlessm4t}
L.~Barrault \emph{et~al.}, ``Seamlessm4t-massively multilingual and multimodal machine translation,'' \emph{arXiv preprint arXiv:2308.11596}, 2023.

\end{thebibliography}

\end{document}